\documentclass[information,article,submit,moreauthors]{Definitions/mdpi} 

\firstpage{1} 
\pubvolume{1}
\issuenum{1}
\articlenumber{0}
\pubyear{2026}
\copyrightyear{2026}
\datereceived{ } 
\daterevised{ }
\dateaccepted{ } 
\datepublished{ } 

\usepackage[UTF8, scheme=plain]{ctex}
\usepackage{placeins}
\usepackage{needspace}
\preto{\abstractkeywords}{\nolinenumbers}
\newcommand{\stopgrad}{\operatorname{stopgrad}}
\definecolor{tsneBlue}{HTML}{4477AA}
\definecolor{tsnePink}{HTML}{EE6677}
\definecolor{tsneGreen}{HTML}{228833}
\definecolor{tsnePurple}{HTML}{AA3377}

\Title{Description-Guided CLIP Adaptation with Local Evidence Calibration for Few-Shot Plant Disease Recognition}

\Author{Zhenkun Ye $^{1}$, Qian Zhou $^{1}$ and Hua Zou $^{1,}$*}

\AuthorNames{Zhenkun Ye, Qian Zhou and Hua Zou}

\address{%
$^{1}$ \quad School of Computer Science, Wuhan University, Wuhan 430072, China; zhenkunye@whu.edu.cn~(Z.Y.); zhouqian@whu.edu.cn~(Q.Z.)}

\corres{Correspondence: zouhua@whu.edu.cn}

\abstract{Accurate plant disease recognition is crucial for intelligent agriculture. However, limited labeled samples in few-shot settings may not adequately cover typical diagnostic symptoms and intra-class variations. Pretrained vision-language models such as CLIP provide visual-semantic knowledge that can complement this limited supervision. Nevertheless, existing CLIP-based few-shot adaptation methods still mainly rely on coarse-grained class semantics and global visual representations. The former lacks explicit disease-specific diagnostic information, while the latter may dilute fine-grained local disease cues. Moreover, excessive task-specific adaptation under few-shot supervision may lead to over-specialization, compromising the generalization ability inherited from large-scale pre-training. To address these limitations, we propose a description-guided two-stage CLIP adaptation method. We introduce symptom and comparative descriptions to enrich class semantics with complementary disease-specific diagnostic information, including typical visible symptoms and discriminative cues between similar classes. The enhanced class semantics further guide the aggregation of class-relevant patch-level evidence to calibrate global image-text matching, supplementing it with fine-grained diagnostic cues. In addition, to balance task adaptation and cross-domain generalization, we split the adaptation process into two stages: (i) adapt the vision-language encoders through lightweight parameter updates to learn task-specific representations and (ii) freeze the adapted encoders and train a linear classifier on top. At test time, the optimized classifier weights are used for in-domain prediction, while class prototypes generated by the adapted text encoder are used for cross-domain prediction to leverage vision-language transferability. Extensive experiments across two backbones demonstrate that the proposed method achieves the highest mean accuracy in most few-shot settings on PlantVillage, Cassava, and Pear, with respective accuracies of 93.16\%, 79.30\%, and 96.56\% under the ViT-L/14 16-shot setting. In class-aligned cross-domain evaluation on PlantDoc, the proposed method achieves 56.96\% accuracy across 21 shared classes, outperforming all compared methods. These results confirm the effectiveness of the proposed method for fine-grained few-shot plant disease recognition and cross-domain generalization.}


\AtBeginEnvironment{thebibliography}{\interlinepenalty=10000}

\begin{document}

\brokenpenalty=10000

\section{Introduction}

Automated plant disease recognition is crucial for crop management and disease control~\citep{pacal2024systematic}. In recent years, advances in deep learning methods and the availability of large-scale annotated datasets have led to substantial progress in plant disease recognition~\citep{long2023classification,osmenaj2025pixels,tanny2025derienet}. However, existing deep learning approaches generally rely on large amounts of labeled data, which are not always readily available in agricultural scenarios~\citep{assimakopoulos2025smart}. Specifically, collecting disease images and obtaining expert annotations are often costly, and some disease categories are rare, making it difficult to obtain sufficient training samples. Therefore, reducing the reliance of plant disease recognition models on large-scale annotated datasets while maintaining reliable classification performance remains an important challenge in agricultural image recognition.

To alleviate this data bottleneck, few-shot learning provides an effective solution~\citep{yang2022survey}. It aims to recognize target classes using only a small number of labeled samples per class~\citep{chen2019closer}. Approaches to few-shot learning include data augmentation, meta-learning, and transfer learning. Data augmentation expands the available training data through sample transformations or synthesis, meta-learning trains models on simulated few-shot tasks to enable rapid adaptation to new tasks, and transfer learning applies knowledge acquired from source tasks to target tasks. By leveraging this prior knowledge, transfer learning can facilitate adaptation when labeled target samples are scarce.

In recent years, few-shot plant disease classification methods based on transfer learning have typically leveraged visual representations learned from source-domain datasets and adapted them to target disease classification tasks using limited labeled samples. For example, Argüeso et al. fine-tuned Inception V3 on source classes from PlantVillage using triplet loss to learn plant image embeddings. They then froze the feature extractor and trained a support vector machine (SVM) classifier using a small number of labeled samples from target classes~\citep{argueso2020fewshot}. Uskaner Hepsa{\u{g}} used ImageNet-pretrained models and further refined the feature extractors using large-scale plant datasets such as PlantCLEF2022 to obtain embeddings better suited to plant images. These embeddings were then used to train classifiers for plant disease classification with limited labeled target samples~\citep{hepsag2023efficient}. Nuthalapati and Tunga used an ImageNet-pretrained ResNet-18 to extract image embeddings. They then used a Transformer to model relationships among support samples and refine their embeddings, and classified query images using Mahalanobis distance~\citep{nuthalapati2021multidomain}. However, these approaches still rely on additional training with labeled data from related domains to obtain representations suitable for few-shot plant disease classification.

Vision-language foundation models such as CLIP have demonstrated strong performance in zero-shot and few-shot classification, opening up new directions for few-shot learning~\citep{radford2021clip}. Unlike conventional transfer learning approaches, which typically require additional data from related domains to train or refine visual feature extractors, CLIP has learned generalizable image–text representations through large-scale contrastive pretraining, enabling effective adaptation with only a small number of target-task labeled samples. According to where adaptation primarily takes place, existing few-shot CLIP adaptation methods can be categorized into prompt adaptation, representation and parameter adaptation, and classification decision adaptation. In the prompt space, CoOp learns continuous context prompts to adjust class text representations, while CoCoOp further uses input image information to dynamically adjust the prompts~\citep{zhou2022coop,zhou2022cocoop}. At the representation and parameter levels, CLIP-Adapter adjusts pretrained features through lightweight adapter modules, while CLIP-LoRA achieves parameter-efficient adaptation of CLIP through low-rank updates~\citep{gao2024clipadapter,hu2022lora,zanella2024lowrank}. At the classification decision level, Tip-Adapter constructs a cache-based classifier from the few-shot support set, whereas LP++ explicitly learns a linear classifier whose weights combine trainable vectors with class text features~\citep{zhang2022tipadapter,huang2024lpplusplus}. 2SFS divides the training process into two stages, first learning task-specific representations through lightweight parameter updates and then freezing the encoders to train a linear classifier, with the aim of balancing target-task adaptation and generalization~\citep{farina2025twostage}.

Although CLIP-based few-shot adaptation methods have achieved promising classification performance, their direct application to fine-grained plant disease recognition remains challenging. First, existing methods typically rely on class names and generic templates to construct class semantic representations. These representations have limited capacity to explicitly encode diagnostic information (e.g., lesion color, shape, texture, and spatial distribution) and to capture subtle differences between similar classes~\citep{yu2025vlpaw}. Second, existing methods typically rely on global image representations for classification, whereas discriminative cues for plant diseases are often concentrated in small, localized regions. Healthy tissue and background information may dilute this local disease evidence in the global representation~\citep{jing2024fineclip,shoaib2025leveraging}. Furthermore, adaptation with only a few target samples may cause the model to become overly specialized to the current task distribution, compromising the generalization ability acquired through CLIP pretraining~\citep{farina2025twostage}.

To address these challenges, we propose a Two-Stage Few-Shot adaptation method that integrates Symptom descriptions, Comparative descriptions, and Local evidence calibration (2SFS-SCL). First, we introduce symptom descriptions to characterize the typical visible symptoms of each class and use comparative descriptions to provide discriminative cues between similar classes, thereby constructing class text prototypes that encode both disease diagnostic semantics and fine-grained discriminative information. Guided by the enhanced class text prototypes, we then aggregate class-relevant local evidence from patch-level visual features across multiple layers. The resulting local logits are used to calibrate global image–text matching results through a residual formulation, improving the model's use of local visual cues for plant disease recognition. In addition, we divide the few-shot adaptation process into two stages. In the first stage, we perform lightweight parameter updates to CLIP's image and text encoders to learn vision-language representations tailored to the target disease classification task. In the second stage, we freeze the adapted encoders, initialize a linear classifier with the class text prototypes, and further optimize its decision boundaries using the few-shot support set. At test time, predictions for in-domain samples use the optimized classifier weights, whereas predictions for cross-domain samples use class text prototypes generated by the adapted text encoder. This design enables better adaptation to the target disease classification task while mitigating the adverse effects of excessive adaptation on CLIP's generalization ability.

\enlargethispage{\baselineskip}
The main contributions of this work are summarized as follows:

\begin{itemize}

\item We propose a CLIP adaptation framework for few-shot fine-grained plant disease classification. By enriching class semantics and improving the use of local visual evidence, the framework enhances fine-grained disease recognition under limited supervision.

\item We construct disease-aware class text prototypes by combining symptom descriptions and comparative descriptions to incorporate disease diagnostic information, thereby enhancing the semantic representation of fine-grained disease classes.

\item We design a global–local evidence calibration mechanism guided by enhanced class text prototypes. The mechanism aggregates class-relevant local evidence from patch-level visual features across multiple layers and uses this evidence to calibrate global image–text matching results, supplementing global matching with fine-grained disease cues that may be diluted in global visual representations.

\item We conduct standard few-shot classification experiments on PlantVillage, Cassava, and Pear, and construct cross-domain test splits from PlantDoc with classes aligned to PlantVillage. The proposed method outperforms existing methods in most few-shot settings. With ViT-L/14 under the 16-shot setting, it achieves accuracies of 93.16\%, 79.30\%, and 96.56\% on the three few-shot classification datasets, respectively, and 56.96\% on the PlantDoc cross-domain test split covering all 21 shared classes.

\end{itemize}

\section{Materials and Methods}

\subsection{Dataset description}

To evaluate the effectiveness of the proposed method, we conduct standard few-shot classification experiments on PlantVillage~\citep{hughes2015plantvillage}, Cassava~\citep{mwebaze2019icassava}, and a self-constructed Pear dataset. We also construct cross-domain test splits from PlantDoc~\citep{singh2020plantdoc} using classes shared with PlantVillage to further assess generalization under distribution shift. These datasets cover diverse crop types, disease categories, and imaging conditions, providing a comprehensive benchmark for performance evaluation. In the standard few-shot experiments, the model is adapted using only $K$ labeled images randomly sampled per class from the training split.

\begin{figure}[H]
\centering
\includegraphics[width=\textwidth]{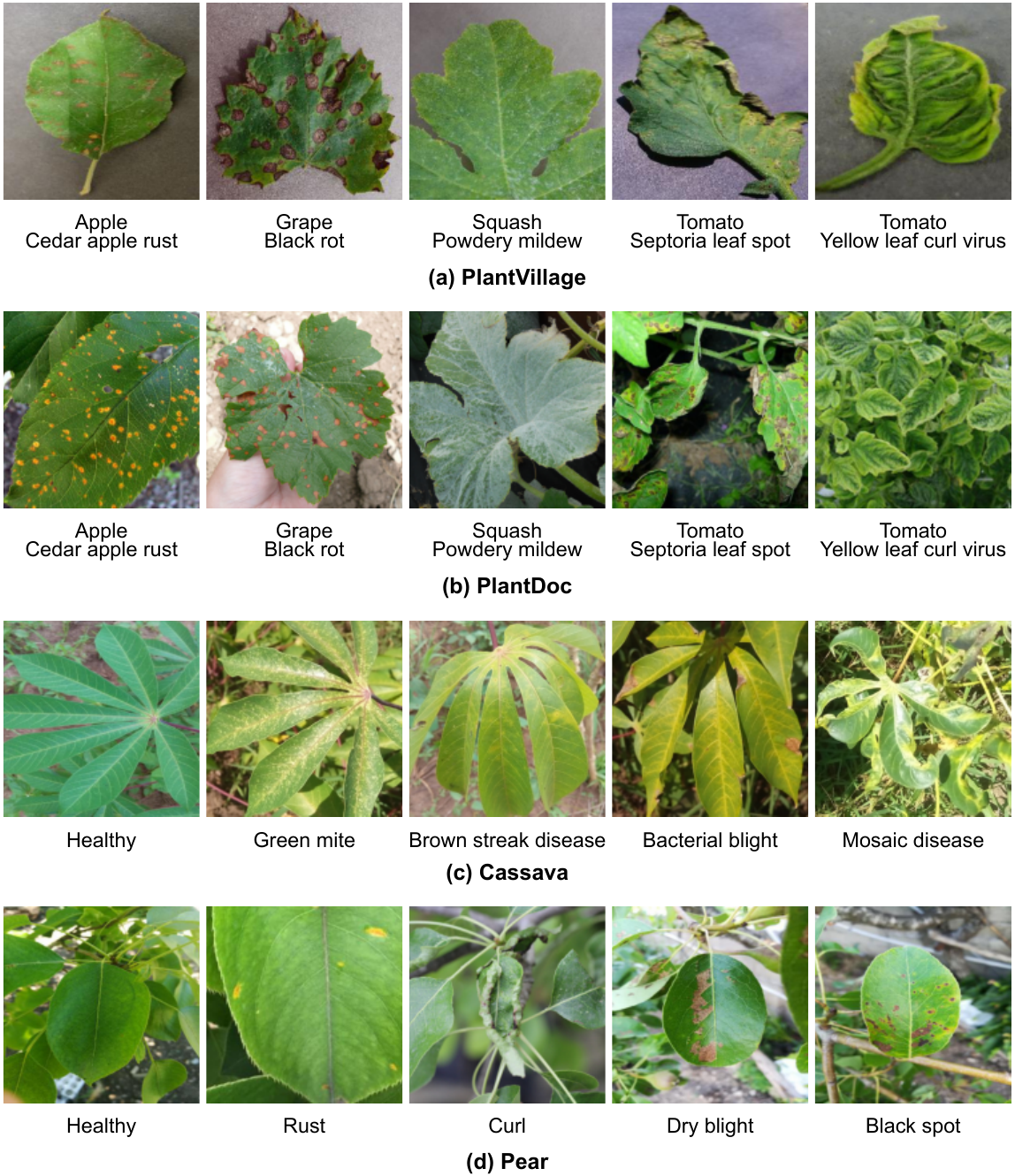}
\caption{Representative images of different datasets. \label{fig:dataset_overview}}
\end{figure}

\subsubsection{PlantVillage}

PlantVillage is a large-scale public benchmark widely used for plant disease classification. It was established to support computer vision research on plant disease recognition by providing open access to images of healthy and diseased plant leaves. The dataset contains 54,305 leaf images from 14 common crops, organized into 38 classes according to crop type and leaf condition, including 12 healthy classes and 26 disease classes. For most crops, classes are further divided according to leaf health status and disease type. The classification task therefore involves both distinguishing among crops and recognizing fine-grained differences in leaf condition within the same crop. Its large size allows adaptation using only a few labeled images per class while retaining ample test samples for performance evaluation.

Images in PlantVillage typically show a single leaf or part of a leaf against a relatively uniform background, with some variation in viewing angle and natural lighting conditions. Disease symptoms are generally clearly visible. Figure~\ref{fig:dataset_overview}(a) shows representative samples from the dataset.

\subsubsection{Cassava}

Cassava is a dataset of field images of cassava leaves released for the iCassava 2019 Fine-Grained Visual Categorization Challenge, focusing on cassava disease classification under natural growing conditions. The dataset contains 5,656 images across five classes: cassava bacterial blight (CBB), cassava brown streak disease (CBSD), cassava green mite (CGM), cassava mosaic disease (CMD), and healthy leaves.

The images were collected in cassava fields across different locations and at different times of day. They include both close-up views of individual leaves and portions of plants containing multiple leaves, with complex backgrounds, varied viewpoints, leaf occlusion, and substantial illumination changes. The extent and severity of symptoms also vary considerably across samples, and some images are out of focus. Figure~\ref{fig:dataset_overview}(c) shows representative samples from the dataset.

\subsubsection{Pear}

Pear is a dataset curated in this study for pear leaf disease and pest classification. Between April and May 2026, we captured 523 pear leaf images in pear orchards in Xinjiang, China, using the rear main camera of an iPhone 16 smartphone (48 MP, f/1.6 aperture). The remaining 1,151 images were sourced from ``Pear Pests and Diseases'' (Version 1)~\citep{roboflowPearV1}, a public dataset hosted on Roboflow Universe and distributed under the Creative Commons Attribution 4.0 International (CC BY 4.0) license.

\begin{figure}[!htbp]
\centering
\includegraphics[width=\textwidth]{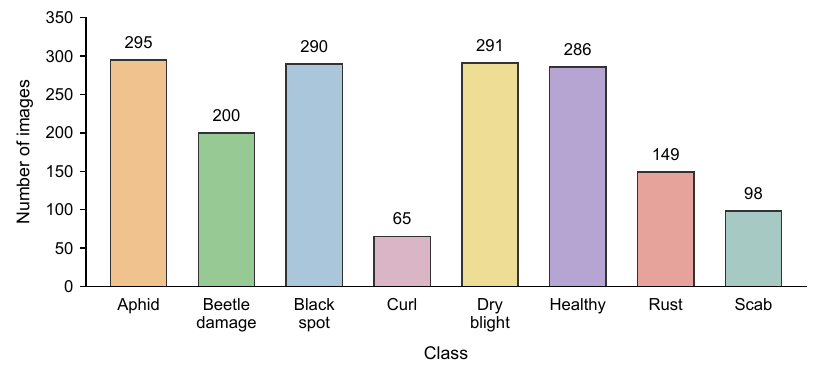}
\caption{Class distribution of the Pear dataset.\label{fig:pear_distribution}}
\end{figure}

We applied a consistent curation procedure to images from both sources. We first excluded images unrelated to pear leaf disease and pest classification, as well as those whose leaf condition could not be reliably assessed because of severe blur, abnormal exposure, or extensive occlusion. Before splitting the dataset, we checked for duplicate and near-duplicate images, manually verified candidate image pairs, and removed confirmed duplicates or near-duplicates. We then standardized class names across sources and assigned image-level labels based on visible disease symptoms, pest-related signs, or healthy leaf appearance. Agricultural experts reviewed these labels, re-examined images with disputed class assignments, and excluded those whose labels remained uncertain. The curated dataset contains 1,674 images across eight classes: black spot, curl, dry blight, rust, scab, aphid, beetle damage, and healthy leaves. The number of images in each class is presented in Figure~\ref{fig:pear_distribution}.

The images include both close-up views of individual leaves and views of branches with multiple leaves, with variations in background, framing, viewing angle, and lighting conditions. Affected leaves exhibit visible symptoms such as dark lesions, yellow-orange spots, localized necrosis, and leaf curling. This variation in disease symptoms, pest-related signs, and imaging conditions provides diverse data for evaluating few-shot classification methods. Figure~\ref{fig:dataset_overview}(d) shows representative samples from the Pear dataset.

\subsubsection{PlantDoc}

PlantDoc is a public dataset of plant disease images captured in natural scenes, developed primarily for plant disease detection and classification research. The dataset contains 2,598 images covering 13 crops and 27 classes, including 17 disease classes and 10 healthy classes. The images exhibit complex backgrounds, varied viewpoints, and substantial variation in illumination. Figure~\ref{fig:dataset_overview}(b) shows representative samples from the PlantDoc dataset.

Using classes shared between PlantDoc and PlantVillage, we constructed three class-aligned cross-domain test splits, denoted as PD-Tomato8, PD-NonTomato13, and PD-Full21. PD-Tomato8 contains eight tomato classes and assesses cross-domain generalization within a single crop, whereas PD-NonTomato13 contains 13 non-tomato classes and assesses cross-domain generalization across multiple crops. PD-Full21 combines the two splits to evaluate overall cross-domain performance on all 21 shared classes. Before constructing the cross-domain test splits, we removed duplicate images appearing in different classes within PlantDoc and checked for near-duplicate images between PlantVillage and PlantDoc. Checks and manual verification identified no duplicate or near-duplicate images between the two datasets. The three final splits contain 712, 1,256, and 1,968 images, respectively.

\subsection{Overview of the Proposed Framework}

\begin{figure}[H]
\centering
\includegraphics[width=\textwidth]{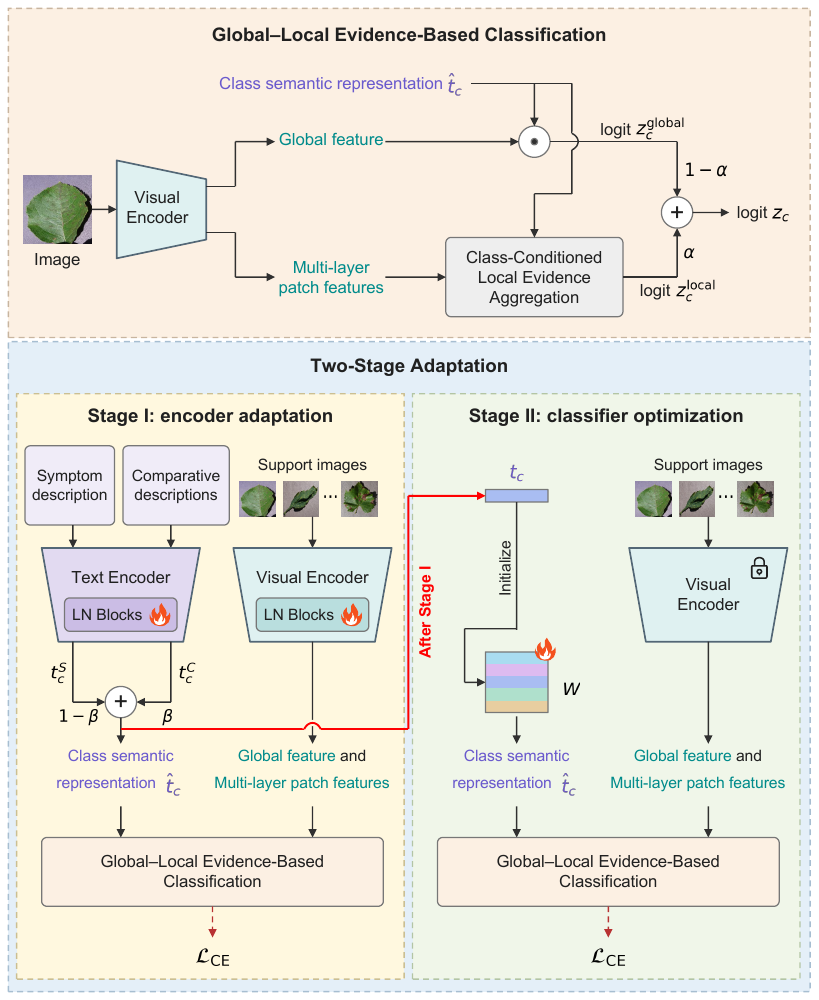}
\caption{Overview of the proposed framework. Both stages share a global--local classification process that uses local logits to calibrate global logits. Stage I constructs class text prototypes from symptom and comparative descriptions and adapts the encoders. Stage II freezes the adapted encoders and optimizes a linear classifier initialized with these prototypes. Class-specific computations are illustrated for class $c$.\label{fig:method_overview}}
\end{figure}

We propose the 2SFS-SCL framework for few-shot plant disease classification, as illustrated in Figure~\ref{fig:method_overview}. The framework combines global and local evidence for classification. The model computes global logits using enhanced class semantic representations and global visual features extracted by the visual encoder. Guided by these class semantic representations, it also aggregates class-relevant evidence from local visual features across multiple encoder layers to obtain local logits. The local logits calibrate the global logits to produce the final classification logits, supplementing global matching with fine-grained disease cues that may be diluted in global representations.

Based on this classification process, we adapt the model in two stages using the few-shot support set~\citep{farina2025twostage}. In Stage I, we use class text prototypes constructed from symptom and comparative descriptions as class semantic representations and perform lightweight parameter updates to the visual and text encoders. These updates adapt the vision-language representations to the target disease classification task. In Stage II, we freeze the adapted encoders and initialize a linear classifier with the class text prototypes obtained in Stage I. We then use its weights as class semantic representations and optimize the classifier using the same classification process.

\subsection{Global--Local Evidence-Based Classification}
\label{sec:global_local_classification}

Because discriminative cues in plant disease images are often localized, their contribution to global visual representations can be diluted by background and healthy tissue information. To address this issue, we develop a global--local evidence-based classification process. Specifically, we aggregate class-relevant local evidence and use the resulting local logits to calibrate global logits through a residual formulation.

\Needspace{12\baselineskip}
\subsubsection{Global Image--Class Matching}

Given an input image \(x\), the visual encoder \(E_I\) extracts a global visual feature \(v=E_I(x)\in\mathbb{R}^d\), where \(d\) is the dimensionality of the shared image--text embedding space. For a task with \(C\) classes, let \(\hat{t}_c\in\mathbb{R}^d\) denote the semantic representation of class \(c\), where \(c=1,\ldots,C\). The global logits are computed from the cosine similarities between the global visual feature and the class semantic representations:
\begin{equation}
z_c^{\mathrm{global}} = \gamma\cos(v,\hat{t}_c)
\end{equation}
where \(\cos(\cdot,\cdot)\) denotes cosine similarity and \(\gamma\) is CLIP's logit scale.

\subsubsection{Class-Conditioned Local Evidence Aggregation}

To aggregate class-relevant local evidence, we first extract patch-level representations from multiple layers of the visual encoder. Let \(\mathcal{L}\) denote the set of visual encoder layers used to compute local evidence. For each selected layer \(l\in\mathcal{L}\), we exclude the classification (CLS) token and sequentially apply the visual encoder's final LayerNorm and visual projection to project the patch tokens into the shared image--text embedding space. We denote the local visual feature corresponding to the \(n\)-th patch by \(v_{l,n}\in\mathbb{R}^d\) and define its relevance to class \(c\) as:
\begin{equation}
r_{l,n,c} = \cos(v_{l,n},\hat{t}_c)
\end{equation}
To assign higher aggregation weights to patches more relevant to class \(c\), we apply softmax normalization to the patch relevance values within each layer, yielding class-conditioned aggregation weights:
\begin{equation}
a_{l,n,c} = \frac{\exp\left(r_{l,n,c}/\tau\right)}{\sum_j\exp\left(r_{l,j,c}/\tau\right)}
\end{equation}
where \(j\) indexes all patch tokens in the current layer and \(\tau\) is the temperature parameter, set to \(0.07\) by default. We compute a weighted sum of the patch relevance values within the layer and multiply it by the logit scale \(\gamma\) to obtain the local logit for class \(c\) at layer \(l\):
\begin{equation}
z_{l,c}^{\mathrm{local}} = \gamma\sum_n a_{l,n,c}r_{l,n,c}
\end{equation}

We then aggregate the local logits for class \(c\) across layers through a weighted sum to obtain its final local logit:
\begin{equation}
z_c^{\mathrm{local}} = \sum_{l\in\mathcal{L}}\omega_l z_{l,c}^{\mathrm{local}}
\end{equation}
where \(\omega_l\) is the fixed aggregation weight for layer \(l\).

\subsubsection{Global--Local Logit Calibration}
\label{sec:global_local_calibration}

The global and local logits of each class are combined through a weighted sum to obtain the final classification logits:
\begin{equation}
\label{eq:global_local_fusion}
z_c = (1-\alpha)z_c^{\mathrm{global}} + \alpha z_c^{\mathrm{local}}
\end{equation}
where \(\alpha\in[0,1]\) is the fusion weight for the local logits, set to \(0.5\) by default. Through this fusion, the local logits supplement global matching with class-relevant local evidence, providing fine-grained disease symptom information that may be diluted in global representations.

\subsection{Stage I: Description-Guided CLIP Adaptation}

In Stage I, we perform lightweight task adaptation of pretrained CLIP using description-guided class text prototypes. For fine-grained plant disease classification, class names alone cannot adequately convey the typical visible symptoms of each class or the subtle differences between similar classes. We therefore construct symptom prototypes and comparative prototypes separately and fuse them in the text embedding space to form class text prototypes. We use these prototypes as class semantic representations and update the LayerNorm parameters in the visual and text encoders on the few-shot support set, following the classification process defined in Section~\ref{sec:global_local_classification}.

\subsubsection{Symptom Prototype Construction}
\label{sec:symptom_prototype}

For each class $c$, we first construct structured class-specific textual information comprising the crop name, disease name, and a description of typical symptoms. Crop and disease names are standardized based on the dataset class labels to reduce the effects of naming variations on text encoding. Symptom descriptions are curated from reputable sources, including the University of Minnesota Extension, the University of California Statewide IPM Program, and published literature~\citep{umnExtension,ucipmGuidelines,sseruwagi2024cassava}. These descriptions summarize visible diagnostic cues such as lesion color and shape, necrotic regions, yellowing patterns, and leaf curling.

We insert this class information into a fixed template, ``a photo of $[\mathrm{crop}_c]$ $[\mathrm{disease}_c]$ leaves with symptoms of $[\mathrm{desc}_c]$.'', to construct the symptom description prompt $p_c^{\mathrm S}$ for class $c$. Here, $[\mathrm{crop}_c]$, $[\mathrm{disease}_c]$, and $[\mathrm{desc}_c]$ denote the crop name, disease name, and description of typical visible symptoms for class $c$, respectively.
We then encode the prompt $p_c^{\mathrm S}$ using the text encoder $E_T$ and normalize the resulting feature to obtain the symptom prototype:
\begin{equation}
t_c^{\mathrm S} = \operatorname{Norm}\left(E_T(p_c^{\mathrm S})\right)
\in\mathbb R^d
\end{equation}
where $\operatorname{Norm}(\cdot)$ denotes $L_2$ normalization. This symptom prototype provides a class semantic reference for the subsequent generation and selection of comparative descriptions.

\subsubsection{Comparative Prototype Construction}

Symptom prototypes capture typical class symptoms but may not adequately represent subtle differences between visually similar disease classes. We therefore compare the target class with similar classes to generate comparative descriptions that highlight its discriminative symptoms and use these descriptions to construct comparative prototypes. This process comprises four steps: similar class retrieval, candidate description generation, candidate description selection, and prototype aggregation~\citep{lee2025comparative}.

First, for each target class $c$, we retrieve the $N_s$ other classes with the highest semantic similarity to it based on cosine similarities between symptom prototypes, forming the similar class set $\mathcal{N}(c)$. We then use a fixed generation prompt template to guide GPT-5.5 in generating comparative descriptions for the target class with reference to these similar classes. The candidate descriptions use specific, concise phrases to describe distinctive visible symptoms of the target class. The generation protocol, prompt template, and representative examples are provided in the Supplementary Materials.

To select candidate descriptions, we insert each candidate comparative description into the class prompt template in Section~\ref{sec:symptom_prototype}, combining the crop name, disease name, and candidate description. Let $p_{c,r}^{\mathrm C}$ denote the $r$-th candidate prompt for class $c$. We encode it with the text encoder $E_T$ to obtain its text feature:
\begin{equation}
t_{c,r}^{\mathrm C} = \operatorname{Norm}\left(E_T(p_{c,r}^{\mathrm C})\right)
\in\mathbb R^d
\end{equation}
Candidate descriptions should be relevant to the target class's symptoms and distinguish it from similar classes. We therefore define the selection score as the sum of textual relevance and an inter-class discriminative margin:
\begin{equation}
\begin{aligned}
q_{c,r} = \cos(t_c^{\mathrm S},t_{c,r}^{\mathrm C})
+ \left[
    \cos(t_c^{\mathrm S},t_{c,r}^{\mathrm C})
    - \max_{k\in\mathcal{N}(c)}\cos(t_k^{\mathrm S},t_{c,r}^{\mathrm C})
  \right]
\end{aligned}
\label{eq:csd_filter_score}
\end{equation}
The first term measures the relevance of the candidate description to the target class's symptom description, while the second measures its relevance margin over the most competitive similar class. We rank candidate descriptions by the score in Equation~\eqref{eq:csd_filter_score} and retain the top $M$ for each class, with $M=5$ by default.

To keep description priors consistent across experimental settings, candidate descriptions are generated and selected offline before model adaptation. For each dataset--backbone combination, the same set of top-$M$ descriptions is used across different shot settings and their corresponding support sets. We aggregate the text features of these fixed descriptions to form the comparative prototype:
\begin{equation}
t_c^{\mathrm C} = \operatorname{Norm}\left(
    \frac{1}{M}\sum_{m=1}^{M}t_{c,m}^{\mathrm C}
\right)
\end{equation}

\subsubsection{Prototype Fusion}

We fuse the symptom and comparative prototypes in the text embedding space to obtain the class text prototype:
\begin{equation}
t_c = (1-\beta)t_c^{\mathrm S} + \beta t_c^{\mathrm C}
\in\mathbb R^d
\end{equation}
where $\beta$ is the fusion weight for the comparative prototype, set to $0.5$ by default. The fused class text prototype $t_c$ encodes both typical symptoms of the class and discriminative cues for distinguishing similar classes, serving as the class semantic representation for the global--local evidence-based classification process in Stage I.

\subsubsection{LayerNorm-Based CLIP Adaptation}

In Stage I, we initialize the model with pretrained CLIP parameters and update only the scale and bias parameters of all LayerNorm layers in the visual and text encoders~\citep{ba2016layernorm}, keeping the remaining backbone parameters frozen.

The symptom descriptions and selected comparative descriptions remain fixed during training. At each training iteration, we re-encode these descriptions with the current text encoder to construct the class text prototypes \(t_c\) through the aggregation and fusion procedures described above. These prototypes serve as the class semantic representations, with \(\hat{t}_c=t_c\). Meanwhile, the current visual encoder extracts global and multi-layer local visual features from the few-shot support images. Based on these class semantic representations and visual features, we compute the final classification logits through the global--local evidence-based classification process described in Section~\ref{sec:global_local_classification}. We then compute the training loss using these logits and the ground-truth class labels to update the LayerNorm parameters.

\subsection{Stage II: Prototype-Initialized Classifier Optimization}

In Stage II, we freeze the vision-language encoders adapted in Stage I and optimize only a bias-free linear classifier initialized with the class text prototypes. Let $t_c^{\mathrm I}$ denote the class text prototype obtained at the end of Stage I. The classifier weight matrix $W\in\mathbb{R}^{C\times d}$ is initialized as:
\begin{equation}
W^{(0)} = \left[
    \operatorname{Norm}(t_1^{\mathrm I}), \ldots,
    \operatorname{Norm}(t_C^{\mathrm I})
\right]^\top
\end{equation}
where the $c$-th row $w_c^\top$ of $W$ corresponds to class $c$.

During optimization, we use the classifier weights $w_c$ as class semantic representations, with $\hat{t}_c=w_c$. Meanwhile, the frozen visual encoder extracts global and multi-layer local visual features from the few-shot support images. Based on these class semantic representations and visual features, we compute the final classification logits for each class through the same classification process described in Section~\ref{sec:global_local_classification}. We then compute the training loss using these logits and the ground-truth class labels to update only the classifier weights $W$.

\subsection{Training Objective and Inference}

In both adaptation stages, we use the same training objective on the few-shot support set and optimize the respective trainable parameters. Given a training image \(x\) with ground-truth class label \(y\), we compute the global and local logits for \(x\) across all classes as described in Section~\ref{sec:global_local_classification}. During training, we define a residual as the difference between the local logits and the global logits with gradients stopped. We then add this weighted residual to the global logits to obtain the final classification logits \(\{z_c\}_{c=1}^{C}\):
\begin{equation}
\label{eq:residual_training}
z_c = z_c^{\mathrm{global}}
    + \alpha\left[z_c^{\mathrm{local}} - \stopgrad\!\left(z_c^{\mathrm{global}}\right)\right],
\qquad c=1,\ldots,C
\end{equation}
where \(\stopgrad(\cdot)\) denotes the stop-gradient operation. This residual formulation produces the same forward output as the weighted fusion in Section~\ref{sec:global_local_calibration}, but preserves the gradient through the global-logit main path during backpropagation to maintain training stability.

We apply softmax to obtain the class probabilities for image \(x\):
\begin{equation}
\label{eq:class_probability}
P(c\mid x) = \frac{\exp(z_c)}{\sum_{j=1}^{C}\exp(z_j)},
\qquad c=1,\ldots,C
\end{equation}
In both stages, we compute the cross-entropy loss using the probability assigned to the ground-truth class \(y\):
\begin{equation}
\label{eq:cross_entropy}
\mathcal{L}_{\mathrm{CE}} = -\log P(y\mid x)
\end{equation}

At test time, both the adapted vision-language encoders and the optimized linear classifier remain frozen. For in-domain testing, we use the class weights optimized in Stage II as class semantic representations, with \(\hat t_c=w_c\). For cross-domain testing, we use the class text prototypes obtained at the end of Stage I, with \(\hat t_c=t_c^{\mathrm I}\). Meanwhile, the adapted visual encoder extracts visual features from input image \(x\). Based on the corresponding class semantic representations and visual features, we compute the final classification logits for each class as described in Section~\ref{sec:global_local_classification}. We then predict the class with the highest logit.

\begingroup
\captionsetup[figure]{skip=4pt}
\captionsetup[table]{skip=4pt}
\makeatletter
\setlength{\@fptop}{0pt}
\setlength{\@fpsep}{8pt}
\setlength{\@fpbot}{0pt plus 1fil}
\makeatother
\section{Results}

\subsection{Experimental Settings}
\label{sec:experimental_settings}

\noindent\textbf{Evaluation Protocol.} We conduct few-shot classification experiments across three datasets (PlantVillage, Cassava, and Pear) and two pretrained CLIP configurations (ViT-B/16 and ViT-L/14)~\citep{radford2021clip,dosovitskiy2021vit}. After removing duplicate and near-duplicate images, we split each dataset into training, validation, and test sets in a 6:2:2 ratio, stratified by class. Following the evaluation protocol of LP++~\citep{huang2024lpplusplus}, we construct each support set by randomly sampling $K$ labeled images per class from the training set, with $K\in\{1,2,4,8,16\}$. This sampling is repeated to obtain 10 support sets for each dataset and shot setting so that evaluation results are less sensitive to sampling variability. In addition, we evaluate cross-domain generalization on PlantDoc. For each test split, we perform 16-shot adaptation on the class-aligned PlantVillage subset and then evaluate the adapted model directly on that split.

\noindent\textbf{Implementation Details.} All experiments are implemented in PyTorch and conducted on an NVIDIA GeForce RTX 4090 GPU with 24 GB of memory. In 2SFS-SCL, training images are augmented with random cropping and random horizontal flipping. All images are then resized to $224\times224$ pixels and normalized using CLIP's mean and standard deviation. Local evidence is extracted from four visual encoder layers: 6, 8, 10, and 12 for ViT-B/16, and 12, 16, 20, and 24 for ViT-L/14. For both backbones, the aggregation weights are fixed at $0.1$, $0.2$, $0.3$, and $0.4$ from shallow to deep. The model is trained using AdamW~\citep{loshchilov2019decoupled} with a batch size of 32. The learning rate is initialized at $2\times10^{-4}$ and follows a cosine annealing schedule implemented with CosineAnnealingLR. Each adaptation run comprises $300\times K$ optimization steps, with $60\%$ allocated to Stage I and $40\%$ to Stage II. Details of the runtime environment and main training settings are provided in Table~\ref{tab:implementation}.

\begin{table}[H]
\caption{Runtime environment and training settings for 2SFS-SCL.\label{tab:implementation}}
\centering
\renewcommand{\arraystretch}{1.10}
\begin{tabularx}{\textwidth}{@{}
>{\hsize=0.7\hsize\linewidth=\hsize\raggedright\arraybackslash}X
>{\hsize=1.3\hsize\linewidth=\hsize\raggedright\arraybackslash}X
@{}}
\toprule
\textbf{Setting} & \textbf{Configuration} \\
\midrule
GPU & NVIDIA GeForce RTX 4090 (24 GB) \\
Operating System & Ubuntu 20.04.2 LTS \\
Python Version & 3.10.20 \\
PyTorch Version & 2.6.0 (CUDA 12.4) \\
\addlinespace[2pt]
Input Image Size & $224\times224$ pixels \\
Optimizer & AdamW \\
Initial Learning Rate & $2\times10^{-4}$ \\
Learning-Rate Scheduler & CosineAnnealingLR, $\eta_{\min}=1\times10^{-6}$ \\
Batch Size & 32 \\
Total Optimization Steps & $300\times K$ \\
Step Allocation & Stage I: 60\%, Stage II: 40\% \\
\bottomrule
\end{tabularx}
\end{table}

\noindent\textbf{Evaluation Metrics.} We evaluate classification performance using Accuracy, Macro-Precision, Macro-Recall, and Macro-F1. Accuracy reflects the overall rate of correct classification. The three macro-averaged metrics are computed as the unweighted means of per-class Precision, Recall, and F1-score, respectively, to assess performance under class imbalance. For a task with $C$ classes and $N$ test samples, including $N_{\mathrm{correct}}$ correctly classified samples, the four metrics are defined as:
\begin{align}
\mathrm{Accuracy} &= \frac{N_{\mathrm{correct}}}{N} \\[4pt]
\mathrm{Macro\text{-}Precision} &= \frac{1}{C}\sum_{c=1}^{C}\frac{TP_c}{TP_c+FP_c} \\[4pt]
\mathrm{Macro\text{-}Recall} &= \frac{1}{C}\sum_{c=1}^{C}\frac{TP_c}{TP_c+FN_c} \\[4pt]
\mathrm{Macro\text{-}F1} &= \frac{1}{C}\sum_{c=1}^{C}\frac{2TP_c}{2TP_c+FP_c+FN_c}
\end{align}
where $TP_c$, $FP_c$, and $FN_c$ denote the one-vs-rest counts of true positives, false positives, and false negatives for class $c$, respectively.

\subsection{Comparison with Others}

We compare 2SFS-SCL with zero-shot CLIP and several representative few-shot CLIP adaptation methods. The few-shot methods include prompt-learning, feature-adaptation, and classifier-adaptation methods, as well as the two-stage method 2SFS:

\begin{itemize}
    \item \textbf{Prompt-learning methods.} CoOp (ctx=16) learns 16 continuous context tokens shared across classes, while CoCoOp generates instance-conditioned context prompts based on the input image~\citep{zhou2022coop,zhou2022cocoop}. PLOT++ learns multiple visual and textual prompts and uses optimal transport to align the resulting image and text representations~\citep{chen2023plot}. KgCoOp regularizes prompt learning with knowledge from pretrained CLIP to mitigate the adverse effects of task adaptation on generalization~\citep{yao2023kgcoop}.

    \item \textbf{Feature-adaptation methods.} CLIP-Adapter optimizes a lightweight residual adapter with a few labeled samples to adapt pretrained CLIP visual features to the target task~\citep{gao2024clipadapter}.

    \item \textbf{Classifier-adaptation methods.} Tip-Adapter-F constructs a cache-based classifier from the few-shot support set and further fine-tunes the cache keys~\citep{zhang2022tipadapter}. TaskRes adds learned task-specific residuals to the zero-shot text classifier weights to adjust class decision boundaries~\citep{yu2023taskres}. LP++ optimizes a linear classifier in the frozen CLIP feature space, incorporating text knowledge~\citep{huang2024lpplusplus}.

    \item \textbf{2SFS} first learns task-specific vision-language representations through parameter-efficient fine-tuning, then freezes the adapted encoders and trains a linear classifier~\citep{farina2025twostage}.
\end{itemize}

We reproduce the compared few-shot adaptation methods using their publicly available implementations and configure the optimizer, learning rate, batch size, and training budget separately for each method. All methods are compared under the backbone settings described in Section~\ref{sec:experimental_settings} and share the same data splits, input size, standard CLIP test preprocessing, and evaluation metrics. For each dataset and shot setting, the few-shot methods are adapted separately on the same 10 sampled support sets and then evaluated on the same test set, with results reported as the mean and sample standard deviation across the 10 evaluations. Zero-shot CLIP does not depend on support sets and is evaluated once on the same test set, so no standard deviation is reported.

\begingroup
\setlength{\intextsep}{6pt plus 1pt minus 1pt}
\setlength{\textfloatsep}{6pt plus 1pt minus 1pt}
\setlength{\floatsep}{8pt plus 1pt minus 1pt}
\subsubsection{Few-Shot Classification Results}

We compare the few-shot classification performance of different methods on PlantVillage, Cassava, and Pear. For both ViT-B/16 and ViT-L/14, we examine classification accuracy across different shot settings and further compare macro-averaged metrics under the 16-shot setting.

\par\addvspace{4pt}\noindent\textbf{PlantVillage}\par\nobreak\vspace{2pt}

As shown in Figure~\ref{fig:few_shot_accuracy_pv}, 2SFS-SCL achieves the highest accuracy on PlantVillage across all shot settings with both backbones, with a particularly pronounced lead at lower shot settings. At 16 shots, 2SFS-SCL also achieves the best performance on all four metrics with both backbones, as reported in Table~\ref{tab:main_pv}. With ViT-B/16, its accuracy and Macro-F1 are 89.45\% and 87.56\%, respectively. The corresponding values for ViT-L/14 reach 93.16\% and 91.80\%, representing respective gains of 1.75 and 2.12 percentage points over the second-best method, 2SFS. Together, these results demonstrate improvements in both overall classification accuracy and average per-class performance under limited supervision.

\begin{figure}[!htbp]
\centering
\includegraphics[width=\textwidth,trim=0bp 9bp 0bp 3bp,clip]{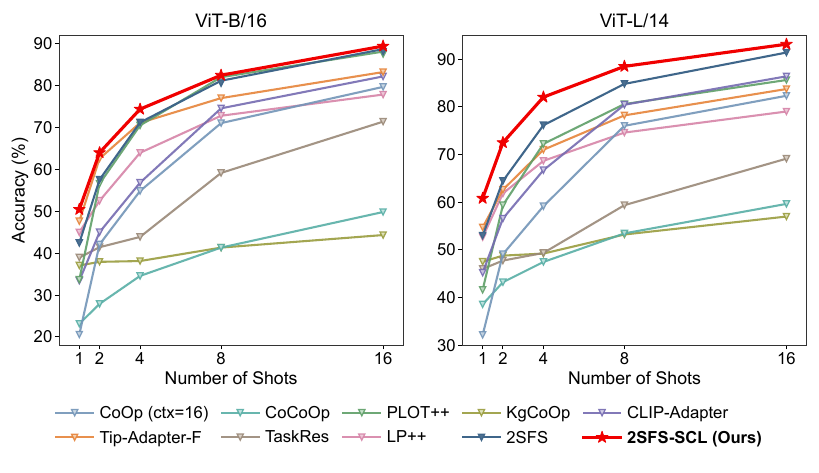}
\caption{Few-shot classification accuracy on PlantVillage with ViT-B/16 and ViT-L/14, averaged over 10 sampled support sets.\label{fig:few_shot_accuracy_pv}}
\end{figure}

\begin{table}[!htbp]
\vspace{24pt}
\centering
\caption{16-shot classification results on PlantVillage (\%). ``CLIP'' refers to zero-shot performance. Results for few-shot adaptation methods are reported as mean \(\pm\) sample standard deviation over 10 sampled support sets. The best results are highlighted in \textbf{bold} and the second-best are \underline{underlined}.\label{tab:main_pv}}
\small
\setlength{\tabcolsep}{4.2pt}
\begin{tabular*}{\textwidth}{@{\extracolsep{\fill}}llcccc@{}}
\toprule
Backbone & Method & Accuracy & Macro-Precision & Macro-Recall & Macro-F1 \\
\midrule
\multirow{11}{*}{ViT-B/16}
& CLIP & 10.12 & 16.54 & 8.26 & 4.64 \\
& CoOp (ctx=16) & 79.71 \(\pm\) 1.89 & 78.74 \(\pm\) 1.20 & 81.23 \(\pm\) 1.88 & 77.48 \(\pm\) 1.69 \\
& CoCoOp & 49.83 \(\pm\) 6.03 & 47.55 \(\pm\) 4.40 & 50.45 \(\pm\) 4.12 & 44.27 \(\pm\) 5.04 \\
& PLOT++ & 88.14 \(\pm\) 1.01 & 85.13 \(\pm\) 0.76 & 89.32 \(\pm\) 0.73 & 85.75 \(\pm\) 0.73 \\
& KgCoOp & 44.33 \(\pm\) 2.44 & 38.55 \(\pm\) 3.23 & 41.71 \(\pm\) 2.95 & 35.68 \(\pm\) 3.10 \\
& CLIP-Adapter & 82.21 \(\pm\) 1.43 & 79.88 \(\pm\) 1.02 & 83.22 \(\pm\) 1.07 & 79.63 \(\pm\) 1.24 \\
& Tip-Adapter-F & 83.25 \(\pm\) 0.82 & 81.62 \(\pm\) 0.77 & 84.38 \(\pm\) 0.72 & 81.16 \(\pm\) 0.72 \\
& TaskRes & 71.41 \(\pm\) 0.83 & 67.98 \(\pm\) 0.72 & 71.23 \(\pm\) 0.73 & 66.30 \(\pm\) 0.82 \\
& LP++ & 77.88 \(\pm\) 1.27 & 76.15 \(\pm\) 0.62 & 79.04 \(\pm\) 0.81 & 74.62 \(\pm\) 0.65 \\
& 2SFS & \underline{88.73} \(\pm\) 0.43 & \underline{85.78} \(\pm\) 0.35 & \underline{89.52} \(\pm\) 0.49 & \underline{86.65} \(\pm\) 0.39 \\
& 2SFS-SCL & \textbf{89.45} \(\pm\) 0.41 & \textbf{86.57} \(\pm\) 0.37 & \textbf{90.57} \(\pm\) 0.35 & \textbf{87.56} \(\pm\) 0.40 \\
\midrule
\multirow{11}{*}{ViT-L/14}
& CLIP & 33.07 & 26.96 & 18.68 & 14.15 \\
& CoOp (ctx=16) & 82.34 \(\pm\) 2.10 & 80.81 \(\pm\) 1.62 & 84.36 \(\pm\) 1.54 & 80.28 \(\pm\) 1.80 \\
& CoCoOp & 59.66 \(\pm\) 2.82 & 58.09 \(\pm\) 2.21 & 62.30 \(\pm\) 2.71 & 55.39 \(\pm\) 2.64 \\
& PLOT++ & 85.65 \(\pm\) 1.02 & 83.92 \(\pm\) 1.52 & 86.92 \(\pm\) 1.56 & 83.74 \(\pm\) 1.92 \\
& KgCoOp & 57.01 \(\pm\) 2.46 & 52.59 \(\pm\) 1.99 & 54.75 \(\pm\) 2.77 & 49.32 \(\pm\) 2.87 \\
& CLIP-Adapter & 86.42 \(\pm\) 0.81 & 84.19 \(\pm\) 0.64 & 87.10 \(\pm\) 0.75 & 84.17 \(\pm\) 0.67 \\
& Tip-Adapter-F & 83.76 \(\pm\) 1.21 & 82.59 \(\pm\) 0.55 & 84.49 \(\pm\) 1.14 & 81.23 \(\pm\) 1.23 \\
& TaskRes & 69.15 \(\pm\) 0.81 & 66.37 \(\pm\) 0.84 & 68.50 \(\pm\) 0.55 & 62.88 \(\pm\) 0.64 \\
& LP++ & 79.04 \(\pm\) 1.72 & 78.64 \(\pm\) 0.93 & 80.11 \(\pm\) 1.16 & 75.98 \(\pm\) 1.37 \\
& 2SFS & \underline{91.41} \(\pm\) 0.59 & \underline{88.95} \(\pm\) 0.68 & \underline{92.01} \(\pm\) 0.48 & \underline{89.68} \(\pm\) 0.70 \\
& 2SFS-SCL & \textbf{93.16} \(\pm\) 0.57 & \textbf{90.93} \(\pm\) 0.72 & \textbf{93.50} \(\pm\) 0.44 & \textbf{91.80} \(\pm\) 0.64 \\
\bottomrule
\end{tabular*}
\end{table}

\Needspace{14\baselineskip}
\par\addvspace{4pt}\noindent\textbf{Cassava}\par\nobreak\vspace{2pt}
\enlargethispage{2\baselineskip}
Figure~\ref{fig:few_shot_accuracy_cassava} shows that 2SFS-SCL achieves the highest accuracy on Cassava from 2 to 16 shots with ViT-B/16 and across all shot settings with ViT-L/14. The 16-shot results in Table~\ref{tab:main_cassava} further show that 2SFS-SCL achieves the best performance on all three macro-averaged metrics with both backbones. With ViT-B/16, its accuracy and Macro-F1 reach 74.18\% and 70.55\%, respectively, exceeding those of 2SFS by 2.38 and 2.33 percentage points. The corresponding values for ViT-L/14 reach 79.30\% and 74.84\%, representing respective gains of 1.43 and 0.87 percentage points over 2SFS. These results indicate that 2SFS-SCL improves few-shot classification not only on leaf images with relatively uniform backgrounds but also on field images with more complex backgrounds and imaging conditions.

\begin{figure}[!htbp]
\centering
\includegraphics[width=\textwidth,trim=0bp 9bp 0bp 3bp,clip]{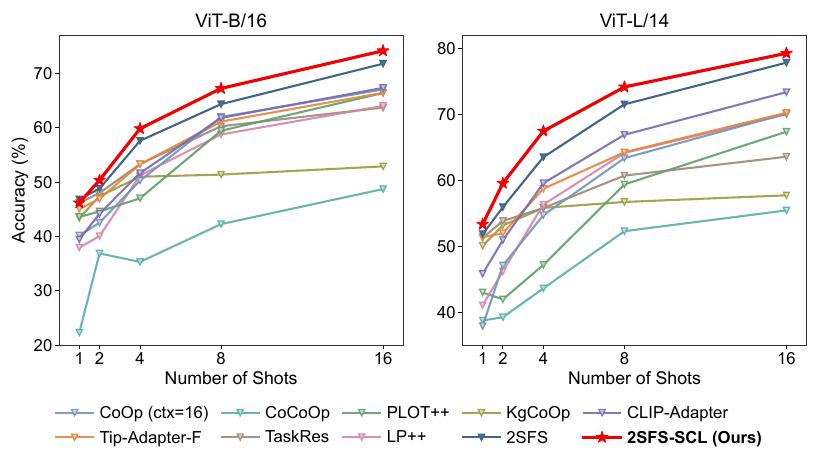}
\caption{Few-shot classification accuracy on Cassava with ViT-B/16 and ViT-L/14, averaged over 10 sampled support sets.\label{fig:few_shot_accuracy_cassava}}
\end{figure}

\begin{table}[!htbp]
\centering
\caption{16-shot classification results on Cassava (\%). ``CLIP'' refers to zero-shot performance. Results for few-shot adaptation methods are reported as mean \(\pm\) sample standard deviation over 10 sampled support sets. The best results are highlighted in \textbf{bold} and the second-best are \underline{underlined}.\label{tab:main_cassava}}
\small
\setlength{\tabcolsep}{4.2pt}
\begin{tabular*}{\textwidth}{@{\extracolsep{\fill}}llcccc@{}}
\toprule
Backbone & Method & Accuracy & Macro-Precision & Macro-Recall & Macro-F1 \\
\midrule
\multirow{11}{*}{ViT-B/16}
& CLIP & 26.13 & 38.12 & 25.62 & 15.23 \\
& CoOp (ctx=16) & 67.07 \(\pm\) 2.37 & 62.39 \(\pm\) 2.06 & 65.08 \(\pm\) 1.93 & 62.88 \(\pm\) 1.90 \\
& CoCoOp & 48.74 \(\pm\) 4.06 & 42.74 \(\pm\) 4.13 & 45.73 \(\pm\) 5.43 & 41.49 \(\pm\) 4.35 \\
& PLOT++ & 66.41 \(\pm\) 3.24 & 63.61 \(\pm\) 2.65 & 66.77 \(\pm\) 2.70 & 64.04 \(\pm\) 2.70 \\
& KgCoOp & 52.91 \(\pm\) 1.61 & 44.13 \(\pm\) 1.50 & 48.54 \(\pm\) 1.54 & 44.90 \(\pm\) 1.63 \\
& CLIP-Adapter & 67.36 \(\pm\) 2.67 & 62.75 \(\pm\) 2.22 & 65.30 \(\pm\) 1.57 & 63.34 \(\pm\) 1.95 \\
& Tip-Adapter-F & 66.45 \(\pm\) 1.82 & 59.63 \(\pm\) 2.01 & 64.73 \(\pm\) 2.10 & 61.04 \(\pm\) 1.97 \\
& TaskRes & 63.73 \(\pm\) 2.17 & 54.27 \(\pm\) 1.81 & 57.10 \(\pm\) 0.97 & 55.18 \(\pm\) 1.30 \\
& LP++ & 64.03 \(\pm\) 2.42 & 59.50 \(\pm\) 1.08 & 63.90 \(\pm\) 1.48 & 60.34 \(\pm\) 1.09 \\
& 2SFS & \underline{71.80} \(\pm\) 2.44 & \underline{67.09} \(\pm\) 2.07 & \underline{70.73} \(\pm\) 1.90 & \underline{68.22} \(\pm\) 1.99 \\
& 2SFS-SCL & \textbf{74.18} \(\pm\) 2.56 & \textbf{69.70} \(\pm\) 2.46 & \textbf{72.54} \(\pm\) 2.16 & \textbf{70.55} \(\pm\) 2.44 \\
\midrule
\multirow{11}{*}{ViT-L/14}
& CLIP & 32.53 & 42.50 & 34.95 & 22.09 \\
& CoOp (ctx=16) & 70.13 \(\pm\) 3.21 & 64.85 \(\pm\) 2.93 & 68.54 \(\pm\) 2.62 & 65.75 \(\pm\) 2.46 \\
& CoCoOp & 55.46 \(\pm\) 5.28 & 47.12 \(\pm\) 12.05 & 54.78 \(\pm\) 10.20 & 47.94 \(\pm\) 10.87 \\
& PLOT++ & 67.40 \(\pm\) 2.83 & 61.83 \(\pm\) 3.35 & 66.05 \(\pm\) 2.90 & 62.90 \(\pm\) 3.07 \\
& KgCoOp & 57.75 \(\pm\) 1.99 & 51.82 \(\pm\) 1.91 & 60.24 \(\pm\) 1.53 & 53.82 \(\pm\) 2.09 \\
& CLIP-Adapter & 73.40 \(\pm\) 2.02 & 68.49 \(\pm\) 2.15 & 71.77 \(\pm\) 2.06 & 69.56 \(\pm\) 1.72 \\
& Tip-Adapter-F & 70.27 \(\pm\) 2.04 & 64.48 \(\pm\) 2.46 & 69.10 \(\pm\) 1.89 & 65.82 \(\pm\) 1.76 \\
& TaskRes & 63.60 \(\pm\) 2.20 & 56.05 \(\pm\) 1.06 & 62.02 \(\pm\) 1.40 & 57.49 \(\pm\) 1.36 \\
& LP++ & 70.04 \(\pm\) 1.97 & 63.62 \(\pm\) 1.74 & 69.06 \(\pm\) 1.64 & 65.25 \(\pm\) 1.63 \\
& 2SFS & \underline{77.87} \(\pm\) 2.38 & \underline{73.40} \(\pm\) 2.99 & \underline{75.60} \(\pm\) 1.13 & \underline{73.97} \(\pm\) 1.77 \\
& 2SFS-SCL & \textbf{79.30} \(\pm\) 2.68 & \textbf{74.45} \(\pm\) 3.02 & \textbf{76.20} \(\pm\) 1.55 & \textbf{74.84} \(\pm\) 2.16 \\
\bottomrule
\end{tabular*}
\end{table}

\par\addvspace{10pt}\noindent\textbf{Pear}\par\nobreak\vspace{2pt}

As shown in Figure~\ref{fig:few_shot_accuracy_pear}, 2SFS-SCL achieves the highest accuracy on Pear across all shot settings with both backbones. Its advantage over 2SFS is more pronounced at lower shot settings. Table~\ref{tab:main_pear} shows that 2SFS already achieves 16-shot accuracies of 94.34\% and 95.81\% with ViT-B/16 and ViT-L/14, respectively. 2SFS-SCL further raises these values to 95.87\% and 96.56\%. Beyond accuracy, it also achieves the best performance on all three macro-averaged metrics with both backbones. With ViT-L/14, its Macro-Precision, Macro-Recall, and Macro-F1 exceed those of 2SFS by 0.41, 0.99, and 0.73 percentage points, respectively. These results indicate that 2SFS-SCL can further improve overall and macro-averaged classification performance even when baseline accuracy is already high.

\begin{figure}[!htbp]
\centering
\includegraphics[width=\textwidth,trim=0bp 9bp 0bp 3bp,clip]{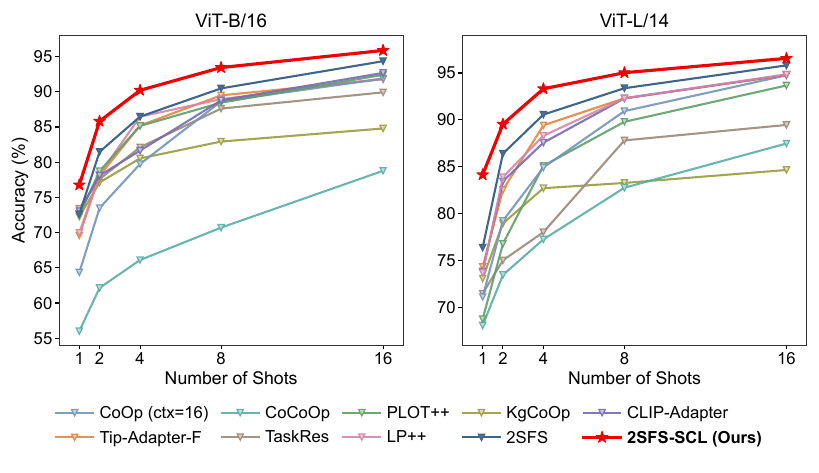}
\caption{Few-shot classification accuracy on Pear with ViT-B/16 and ViT-L/14, averaged over 10 sampled support sets.\label{fig:few_shot_accuracy_pear}}
\end{figure}

\begin{table}[!htbp]
\centering
\caption{16-shot classification results on Pear (\%). ``CLIP'' refers to zero-shot performance. Results for few-shot adaptation methods are reported as mean \(\pm\) sample standard deviation over 10 sampled support sets. The best results are highlighted in \textbf{bold} and the second-best are \underline{underlined}.\label{tab:main_pear}}
\small
\setlength{\tabcolsep}{4.2pt}
\begin{tabular*}{\textwidth}{@{\extracolsep{\fill}}llcccc@{}}
\toprule
Backbone & Method & Accuracy & Macro-Precision & Macro-Recall & Macro-F1 \\
\midrule
\multirow{11}{*}{ViT-B/16}
& CLIP & 38.02 & 33.74 & 30.10 & 24.27 \\
& CoOp (ctx=16) & 91.87 \(\pm\) 1.32 & 91.04 \(\pm\) 1.63 & 92.91 \(\pm\) 1.35 & 91.55 \(\pm\) 1.61 \\
& CoCoOp & 78.78 \(\pm\) 3.53 & 78.49 \(\pm\) 2.51 & 80.28 \(\pm\) 3.77 & 77.77 \(\pm\) 3.41 \\
& PLOT++ & 92.41 \(\pm\) 2.42 & 91.32 \(\pm\) 2.69 & 92.80 \(\pm\) 2.37 & 91.49 \(\pm\) 2.82 \\
& KgCoOp & 84.79 \(\pm\) 1.50 & 83.38 \(\pm\) 1.25 & 85.74 \(\pm\) 1.73 & 83.29 \(\pm\) 1.62 \\
& CLIP-Adapter & 92.68 \(\pm\) 1.02 & 91.57 \(\pm\) 1.38 & 93.25 \(\pm\) 1.37 & 91.98 \(\pm\) 1.41 \\
& Tip-Adapter-F & 91.77 \(\pm\) 1.89 & 90.25 \(\pm\) 2.24 & 92.60 \(\pm\) 2.00 & 90.83 \(\pm\) 2.25 \\
& TaskRes & 89.92 \(\pm\) 1.46 & 88.62 \(\pm\) 1.59 & 91.22 \(\pm\) 1.20 & 89.12 \(\pm\) 1.67 \\
& LP++ & 91.86 \(\pm\) 2.36 & 90.60 \(\pm\) 2.79 & 92.66 \(\pm\) 2.28 & 91.12 \(\pm\) 2.91 \\
& 2SFS & \underline{94.34} \(\pm\) 1.49 & \underline{93.93} \(\pm\) 1.76 & \underline{95.13} \(\pm\) 1.57 & \underline{94.27} \(\pm\) 1.72 \\
& 2SFS-SCL & \textbf{95.87} \(\pm\) 0.99 & \textbf{95.33} \(\pm\) 1.33 & \textbf{96.62} \(\pm\) 1.16 & \textbf{95.78} \(\pm\) 1.20 \\
\midrule
\multirow{11}{*}{ViT-L/14}
& CLIP & 50.90 & 52.18 & 42.46 & 39.00 \\
& CoOp (ctx=16) & 94.75 \(\pm\) 0.89 & 93.87 \(\pm\) 1.16 & 95.01 \(\pm\) 0.86 & 94.20 \(\pm\) 1.14 \\
& CoCoOp & 87.49 \(\pm\) 3.85 & 86.48 \(\pm\) 2.54 & 87.73 \(\pm\) 3.87 & 86.07 \(\pm\) 3.87 \\
& PLOT++ & 93.67 \(\pm\) 1.48 & 91.70 \(\pm\) 1.69 & 93.18 \(\pm\) 1.16 & 91.88 \(\pm\) 1.75 \\
& KgCoOp & 84.67 \(\pm\) 1.77 & 83.77 \(\pm\) 1.50 & 83.59 \(\pm\) 1.50 & 82.56 \(\pm\) 1.86 \\
& CLIP-Adapter & 94.81 \(\pm\) 1.32 & 93.92 \(\pm\) 1.69 & 95.11 \(\pm\) 1.48 & 94.21 \(\pm\) 1.75 \\
& Tip-Adapter-F & 94.76 \(\pm\) 0.83 & 93.39 \(\pm\) 1.03 & 94.85 \(\pm\) 1.03 & 93.78 \(\pm\) 0.96 \\
& TaskRes & 89.47 \(\pm\) 0.97 & 87.91 \(\pm\) 1.16 & 88.39 \(\pm\) 0.82 & 87.69 \(\pm\) 1.14 \\
& LP++ & 94.85 \(\pm\) 1.40 & 93.71 \(\pm\) 1.65 & 94.90 \(\pm\) 1.58 & 94.01 \(\pm\) 1.71 \\
& 2SFS & \underline{95.81} \(\pm\) 1.41 & \underline{95.62} \(\pm\) 1.59 & \underline{95.81} \(\pm\) 1.76 & \underline{95.57} \(\pm\) 1.68 \\
& 2SFS-SCL & \textbf{96.56} \(\pm\) 0.68 & \textbf{96.03} \(\pm\) 0.88 & \textbf{96.80} \(\pm\) 0.70 & \textbf{96.30} \(\pm\) 0.60 \\
\bottomrule
\end{tabular*}
\end{table}

\endgroup
\subsubsection{Cross-Domain Evaluation on PlantDoc}

To evaluate cross-domain generalization, we compare the classification accuracy of different methods on the three class-aligned PlantDoc test splits, as reported in Table~\ref{tab:plantdoc_ood}. On the single-crop test split PD-Tomato8, 2SFS-SCL achieves the highest accuracy with both ViT-B/16 and ViT-L/14, reaching 37.40\% and 40.66\%, respectively. On PD-NonTomato13, which covers multiple crops, it falls below TaskRes and KgCoOp with ViT-B/16 but achieves the highest accuracy of 74.59\% with ViT-L/14. On PD-Full21, which combines the two splits, 2SFS-SCL achieves the highest accuracies of 49.58\% and 56.96\% with the two backbones, exceeding the corresponding second-best results by 1.92 and 2.31 percentage points, respectively. Overall, 2SFS-SCL outperforms the other methods in most cross-domain test settings with both backbones, indicating good generalization under shifts in image distribution.

\begin{table}[H]
\centering
\caption{Class-aligned cross-domain classification accuracy on PlantDoc (\%). ``CLIP'' refers to zero-shot performance. Few-shot adaptation results are reported as mean \(\pm\) sample standard deviation over 10 sampled 16-shot support sets from PlantVillage. The best results are highlighted in \textbf{bold} and the second-best are \underline{underlined}.\label{tab:plantdoc_ood}}
\small
\begin{tabular*}{\textwidth}{@{\extracolsep{\fill}}llccc@{}}
\toprule
Backbone & Method & PD-Tomato8 & PD-NonTomato13 & PD-Full21 \\
\midrule
\multirow{11}{*}{ViT-B/16}
& CLIP & 19.66 & 48.57 & 33.79 \\
& CoOp (ctx=16) & 33.71 \(\pm\) 3.12 & 55.62 \(\pm\) 2.93 & 42.15 \(\pm\) 1.81 \\
& CoCoOp & 27.99 \(\pm\) 2.65 & 62.12 \(\pm\) 2.51 & 45.25 \(\pm\) 2.76 \\
& PLOT++ & 26.40 \(\pm\) 3.86 & 55.05 \(\pm\) 4.54 & 37.40 \(\pm\) 2.59 \\
& KgCoOp & 28.64 \(\pm\) 1.86 & \underline{66.02} \(\pm\) 0.61 & 47.61 \(\pm\) 0.56 \\
& CLIP-Adapter & 32.22 \(\pm\) 4.16 & 56.56 \(\pm\) 3.22 & 43.80 \(\pm\) 1.55 \\
& Tip-Adapter-F & 33.93 \(\pm\) 2.87 & 60.03 \(\pm\) 2.26 & 44.56 \(\pm\) 1.85 \\
& TaskRes & 33.97 \(\pm\) 1.60 & \textbf{66.09} \(\pm\) 0.90 & \underline{47.66} \(\pm\) 1.36 \\
& LP++ & 29.71 \(\pm\) 3.49 & 61.59 \(\pm\) 2.30 & 44.38 \(\pm\) 2.34 \\
& 2SFS & \underline{34.58} \(\pm\) 3.64 & 63.64 \(\pm\) 1.58 & 45.69 \(\pm\) 1.21 \\
& 2SFS-SCL & \textbf{37.40} \(\pm\) 3.90 & 64.24 \(\pm\) 1.79 & \textbf{49.58} \(\pm\) 1.10 \\
\midrule
\multirow{11}{*}{ViT-L/14}
& CLIP & 24.02 & 63.69 & 45.43 \\
& CoOp (ctx=16) & 35.64 \(\pm\) 6.39 & 61.65 \(\pm\) 2.11 & 47.49 \(\pm\) 1.21 \\
& CoCoOp & 31.90 \(\pm\) 4.96 & 68.60 \(\pm\) 3.70 & 51.67 \(\pm\) 2.01 \\
& PLOT++ & 30.44 \(\pm\) 3.84 & 64.19 \(\pm\) 2.81 & 47.77 \(\pm\) 1.49 \\
& KgCoOp & 34.97 \(\pm\) 1.34 & 71.83 \(\pm\) 0.84 & 53.63 \(\pm\) 1.00 \\
& CLIP-Adapter & 36.28 \(\pm\) 2.52 & 66.00 \(\pm\) 2.07 & 50.42 \(\pm\) 1.18 \\
& Tip-Adapter-F & \underline{38.02} \(\pm\) 3.71 & 65.89 \(\pm\) 3.24 & 49.28 \(\pm\) 2.04 \\
& TaskRes & 36.05 \(\pm\) 1.75 & 72.44 \(\pm\) 0.68 & 54.06 \(\pm\) 0.69 \\
& LP++ & 36.62 \(\pm\) 2.61 & 68.19 \(\pm\) 1.98 & 50.62 \(\pm\) 1.71 \\
& 2SFS & 37.09 \(\pm\) 2.22 & \underline{72.71} \(\pm\) 1.53 & \underline{54.65} \(\pm\) 1.08 \\
& 2SFS-SCL & \textbf{40.66} \(\pm\) 3.45 & \textbf{74.59} \(\pm\) 1.38 & \textbf{56.96} \(\pm\) 1.54 \\
\bottomrule
\end{tabular*}
\end{table}

\subsubsection{Efficiency Analysis}

To examine the trade-off between classification performance and inference efficiency, we compare 2SFS-SCL with other methods under the ViT-L/14 16-shot setting. Table~\ref{tab:efficiency_comparison} reports total parameter counts, floating-point operations (FLOPs), inference speeds in frames per second (FPS), and classification accuracies.

\begin{table}[H]
\caption{Efficiency comparison of different methods in terms of parameters, FLOPs, and FPS, along with their classification accuracies on PlantVillage, Cassava, and Pear under the ViT-L/14 16-shot setting.}
\label{tab:efficiency_comparison}
\centering
\resizebox{\textwidth}{!}{%
\begin{tabular}{lcccccc}
\toprule
\multirow[c]{2}{*}[-0.5ex]{Method}
& \multirow[c]{2}{*}[-0.5ex]{Params (M)}
& \multirow[c]{2}{*}[-0.5ex]{FLOPs (G)}
& \multirow[c]{2}{*}[-0.5ex]{FPS}
& \multicolumn{3}{c}{Accuracy (\%)} \\
\cmidrule(lr){5-7}
& & & & PlantVillage & Cassava & Pear \\
\midrule
CoOp (ctx=16)   & 427.63 & 162.45 & 551.5 & 82.34 & 70.13 & 94.75 \\
CoCoOp          & 427.69 & 669.26 & 97.2  & 59.66 & 55.46 & 87.49 \\
Tip-Adapter-F   & 428.08 & 162.45 & 548.8 & 83.76 & 70.27 & 94.76 \\
CLIP-Adapter    & 427.91 & 162.45 & 530.5 & 86.42 & 73.40 & 94.81 \\
PLOT++          & 427.64 & 654.11 & 126.3 & 85.65 & 67.40 & 93.67 \\
KgCoOp          & 427.62 & 162.45 & 549.6 & 57.01 & 57.75 & 84.67 \\
TaskRes         & 427.65 & 162.45 & 535.9 & 69.15 & 63.60 & 89.47 \\
LP++            & 427.65 & 162.45 & 548.9 & 79.04 & 70.04 & 94.85 \\
2SFS            & 427.65 & 162.45 & 530.8 & 91.41 & 77.87 & 95.81 \\
2SFS-SCL        & 427.65 & 164.13 & 507.7 & 93.16 & 79.30 & 96.56 \\
\bottomrule
\end{tabular}%
}
\end{table}

Among the compared methods, CoOp, KgCoOp, LP++, and Tip-Adapter-F all require 162.45 G FLOPs and achieve inference speeds ranging from 548.8 to 551.5 FPS. These methods offer high inference efficiency, but their classification accuracies on all three datasets are lower than those of 2SFS-SCL. In contrast, PLOT++ achieves competitive classification performance on some datasets, but requires 654.11 G FLOPs and runs at 126.3 FPS. CoCoOp incurs an even higher computational cost of 669.26 G FLOPs and runs at only 97.2 FPS, without corresponding gains in classification performance.

2SFS-SCL achieves the highest classification accuracy on all three datasets, with 427.65 M parameters, 164.13 G FLOPs, and an inference speed of 507.7 FPS. Compared with 2SFS, 2SFS-SCL improves classification accuracy on PlantVillage, Cassava, and Pear by 1.75, 1.43, and 0.75 percentage points, respectively. Its parameter count remains unchanged, with FLOPs increasing by only about 1.04\% and inference speed decreasing by about 4.35\%. This additional cost mainly comes from processing patch-level features across multiple layers and aggregating evidence for local evidence calibration, which supplements global matching with fine-grained disease cues. Overall, 2SFS-SCL achieves higher classification performance with limited additional computational cost, striking a good balance between classification accuracy and inference efficiency.

\FloatBarrier
\subsection{Ablation Study}

\subsubsection{Component Ablation}
\label{sec:component_ablation}

To evaluate the effects of description-guided class semantic representations and local evidence calibration, we conduct a component ablation study on PlantVillage with ViT-L/14. The results are reported in Table~\ref{tab:component_ablation}.

\begin{table}[htbp]
\centering
\caption{Component ablation of 2SFS-SCL on PlantVillage using ViT-L/14 (\%). \(S\), \(C\), and \(L\) denote Symptom prototypes, Comparative prototypes, and Local evidence calibration, respectively. Results are reported as mean accuracy \(\pm\) sample standard deviation over 10 sampled support sets, with the best mean accuracy in each shot setting highlighted in \textbf{bold}.}
\label{tab:component_ablation}
\small
\setlength{\tabcolsep}{3pt}
\begin{tabularx}{\textwidth}{ccc@{\hspace{9pt}}CCCCC}
\toprule
$S$ & $C$ & $L$ & 1-shot & 2-shot & 4-shot & 8-shot & 16-shot \\
\midrule
-- & -- & --
& \(53.00 \pm 2.95\) & \(64.47 \pm 1.81\) & \(76.16 \pm 1.40\) & \(84.83 \pm 1.04\) & \(91.41 \pm 0.59\) \\
\(\checkmark\) & -- & --
& \(53.36 \pm 3.15\) & \(66.21 \pm 1.58\) & \(77.38 \pm 1.12\) & \(85.76 \pm 1.00\) & \(91.91 \pm 0.66\) \\
\(\checkmark\) & \(\checkmark\) & --
& \(57.47 \pm 3.44\) & \(69.20 \pm 2.07\) & \(79.09 \pm 1.21\) & \(86.39 \pm 1.06\) & \(91.98 \pm 0.72\) \\
-- & -- & \(\checkmark\)
& \(56.84 \pm 2.80\) & \(67.62 \pm 1.58\) & \(78.83 \pm 1.27\) & \(86.85 \pm 0.94\) & \(92.51 \pm 0.53\) \\
\(\checkmark\) & -- & \(\checkmark\)
& \(56.99 \pm 2.71\) & \(69.88 \pm 1.32\) & \(80.53 \pm 1.09\) & \(87.72 \pm 0.97\) & \(93.02 \pm 0.57\) \\
\(\checkmark\) & \(\checkmark\) & \(\checkmark\)
& \(\mathbf{60.83} \pm 3.04\) & \(\mathbf{72.52} \pm 1.65\) & \(\mathbf{82.06} \pm 1.01\) & \(\mathbf{88.51} \pm 0.97\) & \(\mathbf{93.16} \pm 0.57\) \\
\bottomrule
\end{tabularx}
\end{table}

As shown in Table~\ref{tab:component_ablation}, without local evidence calibration, symptom prototypes improve accuracy by 0.36, 1.74, 1.22, 0.93, and 0.50 percentage points at 1, 2, 4, 8, and 16 shots, respectively. Adding comparative prototypes yields further gains of 4.11, 2.99, 1.71, 0.63, and 0.07 percentage points across the same five shot settings. With local evidence calibration enabled, symptom and comparative descriptions also yield positive gains across all shot settings. These two types of descriptions progressively enrich class semantics with typical symptoms and inter-class discriminative information, providing a more informative basis for distinguishing similar disease classes. In particular, comparative descriptions yield more pronounced additional gains at lower shot settings, highlighting the value of inter-class discriminative information in complementing limited visual supervision.

Local evidence calibration improves classification accuracy across all shot settings when class semantic representations are based on generic class-name prompts, symptom prototypes, or fused class text prototypes. At 1 shot, the gains under these three settings are 3.84, 3.63, and 3.36 percentage points, respectively. At 16 shots, the corresponding gains are 1.10, 1.11, and 1.18 percentage points. Local evidence calibration supplements global matching with class-relevant local evidence, enabling the model to better exploit fine-grained disease cues that may be diluted in global representations and thereby improve few-shot classification performance.

The full 2SFS-SCL model achieves the highest mean accuracy across all shot settings. Compared with 2SFS, it improves accuracy by 7.83, 8.05, 5.90, 3.68, and 1.75 percentage points at 1, 2, 4, 8, and 16 shots, respectively, with more pronounced gains at lower shot settings. Combining symptom and comparative descriptions with local evidence calibration yields higher accuracy than using the descriptions or local evidence calibration alone. This joint advantage indicates complementarity between enhanced class semantic representations and class-relevant local evidence: the descriptions provide a more specific class semantic reference for local evidence aggregation, and local evidence calibration uses the resulting local cues to supplement global matching.

\begingroup
\setlength{\intextsep}{9pt plus 1pt minus 1pt}
\subsubsection{Analysis of Two-Stage Adaptation}
\label{sec:two_stage_adaptation}

To examine the effects of two-stage adaptation on in-domain classification performance and cross-domain generalization, we compare single-stage and two-stage adaptation with the same total number of optimization steps. Single-stage adaptation continuously updates the vision-language encoders to adapt to the target task, whereas two-stage adaptation first updates the encoders to learn task-specific vision-language representations, then freezes them and trains a linear classifier.

\begin{table}[H]
\centering
\caption{In-domain classification accuracy of single-stage and two-stage adaptation on PlantVillage, Cassava, and Pear under the 16-shot setting (\%). Few-shot adaptation results are reported as mean \(\pm\) sample standard deviation over 10 sampled support sets.}
\label{tab:two_stage_source}
\small
\setlength{\tabcolsep}{4pt}
\begin{tabularx}{\textwidth}{l>{\centering\arraybackslash}XCCC}
\toprule
Backbone & Strategy & PlantVillage & Cassava & Pear \\
\midrule
\multirow{3}{*}{ViT-B/16} & Single-stage & \(83.80 \pm 0.65\) & \(74.20 \pm 2.67\) & \(95.48 \pm 0.85\) \\
& Two-stage & \(89.45 \pm 0.41\) & \(74.18 \pm 2.56\) & \(95.87 \pm 0.99\) \\
& $\Delta$ & \textcolor{red}{\(+5.65\)} & \textcolor{green!50!black}{\(-0.02\)} & \textcolor{red}{\(+0.39\)} \\
\midrule
\multirow{3}{*}{ViT-L/14} & Single-stage & \(90.91 \pm 0.57\) & \(79.55 \pm 2.36\) & \(96.59 \pm 0.63\) \\
& Two-stage & \(93.16 \pm 0.57\) & \(79.30 \pm 2.68\) & \(96.56 \pm 0.68\) \\
& $\Delta$ & \textcolor{red}{\(+2.25\)} & \textcolor{green!50!black}{\(-0.25\)} & \textcolor{green!50!black}{\(-0.03\)} \\
\bottomrule
\end{tabularx}
\end{table}

For in-domain evaluation, single-stage adaptation uses the final class text prototypes as class semantic representations, whereas two-stage adaptation uses the classifier weights optimized in Stage II. Table~\ref{tab:two_stage_source} compares the 16-shot classification results of the two strategies on PlantVillage, Cassava, and Pear. On PlantVillage, two-stage adaptation improves accuracy by 5.65 and 2.25 percentage points with ViT-B/16 and ViT-L/14, respectively. On Cassava and Pear, the two strategies achieve similar accuracies, with absolute differences no greater than 0.39 percentage points. These results indicate that the model can maintain or even improve in-domain classification performance by shifting subsequent optimization from the encoders to the classifier without changing the total number of optimization steps.

\begin{table}[H]
\centering
\caption{Class-aligned cross-domain classification accuracy of single-stage and two-stage adaptation on PlantDoc (\%). Few-shot adaptation results are reported as mean \(\pm\) sample standard deviation over 10 sampled 16-shot support sets from PlantVillage.}
\label{tab:two_stage_cross_domain}
\small
\setlength{\tabcolsep}{4pt}
\begin{tabularx}{\textwidth}{l>{\centering\arraybackslash}XCCC}
\toprule
Backbone & Strategy & PD-Tomato8 & PD-NonTomato13 & PD-Full21 \\
\midrule
\multirow{3}{*}{ViT-B/16} & Single-stage & \(38.06 \pm 3.93\) & \(63.85 \pm 2.05\) & \(48.49 \pm 1.79\) \\
& Two-stage & \(37.40 \pm 3.90\) & \(64.24 \pm 1.79\) & \(49.58 \pm 1.10\) \\
& $\Delta$ & \textcolor{green!50!black}{\(-0.66\)} & \textcolor{red}{\(+0.39\)} & \textcolor{red}{\(+1.09\)} \\
\midrule
\multirow{3}{*}{ViT-L/14} & Single-stage & \(40.39 \pm 3.39\) & \(73.85 \pm 1.28\) & \(56.28 \pm 1.22\) \\
& Two-stage & \(40.66 \pm 3.45\) & \(74.59 \pm 1.38\) & \(56.96 \pm 1.54\) \\
& $\Delta$ & \textcolor{red}{\(+0.27\)} & \textcolor{red}{\(+0.74\)} & \textcolor{red}{\(+0.68\)} \\
\bottomrule
\end{tabularx}
\end{table}

For cross-domain evaluation, both strategies use class text prototypes \(t_c\) generated by their respective text encoders for inference. Single-stage adaptation uses the vision-language encoders obtained at the end of the full optimization process, whereas two-stage adaptation uses the encoders frozen after Stage I. Table~\ref{tab:two_stage_cross_domain} compares the cross-domain classification results of the two strategies on PlantDoc. The results show that two-stage adaptation achieves higher mean accuracy in five of the six cross-domain test settings. On PD-Full21, which covers all 21 shared classes, two-stage adaptation improves accuracy by 1.09 and 0.68 percentage points with ViT-B/16 and ViT-L/14, respectively.

These comparisons support the two-stage adaptation strategy in terms of both target-task classification and cross-domain generalization. After adapting the representations to the target task in Stage I, this strategy freezes the encoders to preserve the vision-language representations used for cross-domain inference. It then uses few-shot supervision to refine classification decision boundaries in the fixed representation space, balancing task adaptation and cross-domain generalization.

\endgroup

\subsubsection{Parameter Sensitivity}

To examine the effects of fusion weights on classification performance, we analyze global--local evidence fusion and prototype fusion on PlantVillage under the ViT-L/14 16-shot setting. Figure~\ref{fig:param_sensitivity} presents the results for the local evidence fusion weight \(\alpha\) and the comparative prototype fusion weight \(\beta\). The \(\alpha\) sweep fixes \(\beta=0.5\), and the \(\beta\) sweep fixes \(\alpha=0.5\), with all other settings following the default 2SFS-SCL configuration.

\begin{figure}[H]
\centering
\includegraphics[width=\textwidth,trim=0bp 16bp 0bp 8bp,clip]{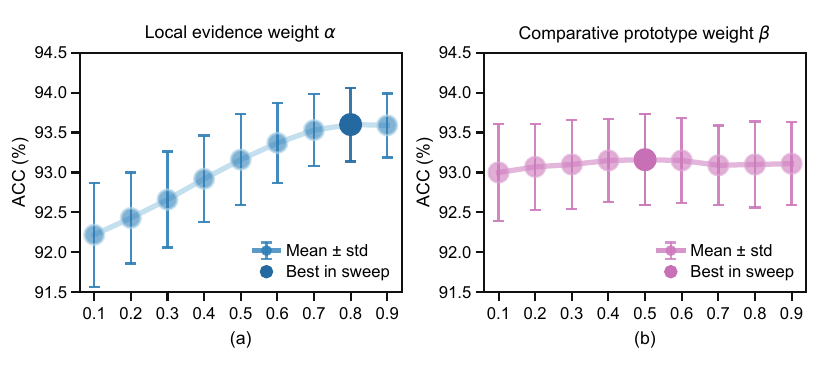}
\caption{Parameter sensitivity of 2SFS-SCL on PlantVillage under the ViT-L/14 16-shot setting. (a) Local evidence fusion weight \(\alpha\), with \(\beta=0.5\); (b) comparative prototype fusion weight \(\beta\), with \(\alpha=0.5\). Results are reported as mean accuracy \(\pm\) sample standard deviation over 10 sampled support sets.\label{fig:param_sensitivity}}
\end{figure}

As shown in Figure~\ref{fig:param_sensitivity}(a), mean accuracy generally increases as \(\alpha\) rises from 0.1 to 0.8, reaching its highest value in this sweep at \(\alpha=0.8\). Mean accuracy remains largely unchanged when the weight is further increased to 0.9. These results indicate that appropriately increasing the weight assigned to local evidence in classification decisions helps improve classification performance by supplementing global matching. Meanwhile, Figure~\ref{fig:param_sensitivity}(b) shows small fluctuations in mean accuracy as \(\beta\) varies from 0.1 to 0.9, with the highest mean accuracy in this sweep achieved at \(\beta=0.5\). These results indicate that the model has low sensitivity to the fusion ratio between symptom and comparative prototypes.

\subsubsection{Layer Aggregation Analysis}

2SFS-SCL aggregates local logits from different visual encoder layers through a weighted sum to calibrate global matching results. To examine the effects of layer counts and aggregation weights on classification performance, we compare different configurations on PlantVillage under the ViT-L/14 16-shot setting, as reported in Table~\ref{tab:local_layer_analysis}. For each layer count, we retain the deepest layers from \(\{12,16,20,24\}\). All other settings follow the default 2SFS-SCL configuration.

\begin{table}[H]
\caption{Classification accuracy for different layer counts and aggregation weights on PlantVillage under the ViT-L/14 16-shot setting (\%). Results are reported as mean \(\pm\) sample standard deviation over 10 sampled support sets.\label{tab:local_layer_analysis}}
\small
\setlength{\tabcolsep}{3.5pt}
\begin{tabularx}{\textwidth}{c>{\centering\arraybackslash}Xc}
\toprule
\textbf{Number of layers} & \textbf{Aggregation weights} & \textbf{Accuracy \(\pm\) Std.}\\
\midrule
0 & -- & 91.98 \(\pm\) 0.72\\
1 & [1.0] & 93.10 \(\pm\) 0.48\\
2 & [0.5, 0.5] & 93.45 \(\pm\) 0.50\\
3 & [1/3, 1/3, 1/3] & 93.23 \(\pm\) 0.55\\
4 & [0.25, 0.25, 0.25, 0.25] & 92.95 \(\pm\) 0.55\\
4 & [0.1, 0.2, 0.3, 0.4] & 93.16 \(\pm\) 0.57\\
\bottomrule
\end{tabularx}
\end{table}

As shown in Table~\ref{tab:local_layer_analysis}, all configurations with local evidence achieve higher accuracy than global matching alone. These results indicate that local evidence effectively complements global matching. Under equal weighting, the two-layer configuration achieves the highest mean accuracy of 93.45\% in this setting. Increasing the number of layers to three or four yields no further improvement. Cross-layer weighting also affects classification performance: for the same four layers, replacing equal weights with fixed weights that increase with depth raises accuracy from 92.95\% to 93.16\%. In the main experiments, we use the same four-layer weighted configuration to draw on local details from shallower layers and semantic information from deeper layers, with progressively higher weights emphasizing deeper-layer evidence during aggregation.

\subsection{Visualization and Analysis}

\subsubsection{Feature Visualization}

To qualitatively examine intra-class compactness and inter-class separation, we use t-SNE~\citep{vandermaaten2008visualizing} to project test-image features into a two-dimensional space. Figure~\ref{fig:image_feature_tsne} compares the feature distributions produced by 2SFS and 2SFS-SCL under the ViT-L/14 16-shot setting. The visualizations show eight tomato classes from PlantVillage, all five classes from Cassava, and all eight classes from Pear. Different colors denote different classes.

\begin{figure}[H]
\centering
\includegraphics[width=\textwidth]{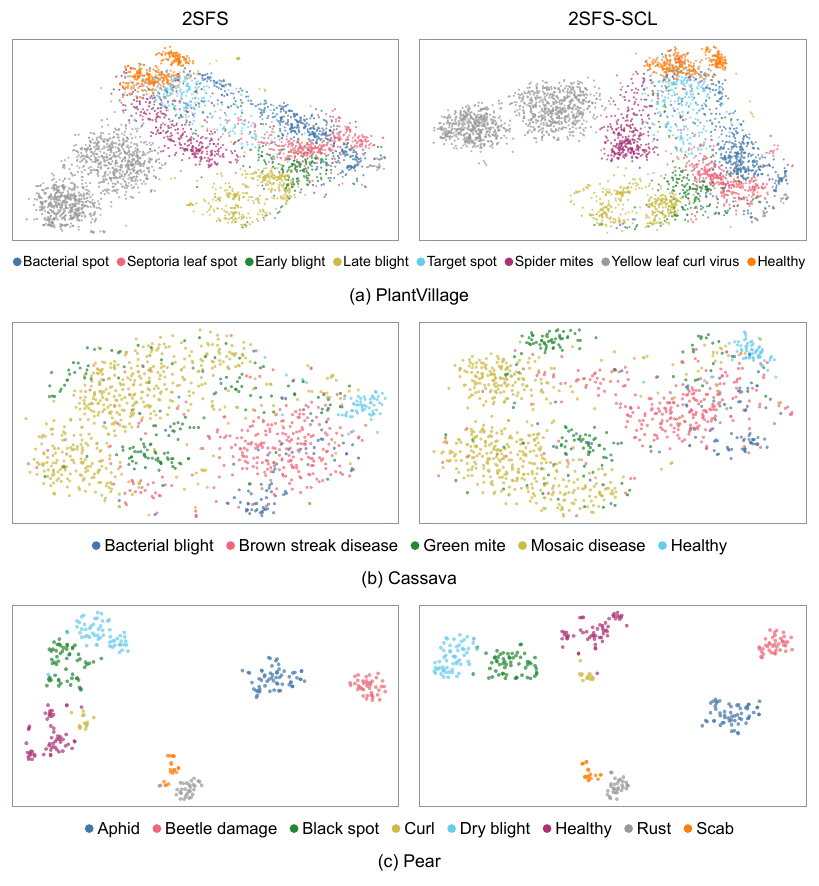}
\caption{t-SNE visualization of test-image features produced by 2SFS and 2SFS-SCL under the ViT-L/14 16-shot setting.\label{fig:image_feature_tsne}}
\end{figure}

As shown in Figure~\ref{fig:image_feature_tsne}, 2SFS-SCL generally exhibits better class separability across the three datasets. On PlantVillage and Cassava, 2SFS-SCL shows more compact clusters for some classes and less inter-class mixing than 2SFS. For example, the main clusters of Bacterial spot (\textcolor{tsneBlue}{\textbf{blue}}) and Septoria leaf spot (\textcolor{tsnePink}{\textbf{pink}}) are better separated on PlantVillage, and some Green mite (\textcolor{tsneGreen}{\textbf{green}}) samples form more distinct local clusters on Cassava. On Pear, both methods produce relatively clear class clusters, with 2SFS-SCL showing greater intra-class compactness for some classes, such as Black spot (\textcolor{tsneGreen}{\textbf{green}}) and \mbox{Healthy (\textcolor{tsnePurple}{\textbf{magenta}})}. Overall, the visualizations show improved intra-class compactness and inter-class separation, consistent with the improvements in classification performance reported above.

\subsubsection{Local Evidence Visualization}

The Grad-CAM~\citep{selvaraju2017gradcam} heatmaps in Figure~\ref{fig:patch_class_relevance} show changes in the regions the model focuses on as components are introduced. The baseline 2SFS generally exhibits dispersed class-relevant responses across both leaf and background regions. With symptom and comparative prototypes, scattered background responses decrease, while local responses within the leaves become more prominent. When local evidence calibration is further introduced, the main response regions within the leaves are retained, and some background responses are further reduced. These changes help the model focus on class-relevant local regions, providing more targeted visual evidence for fine-grained plant disease classification.

\begin{figure}[H]
\centering
\includegraphics[width=\textwidth]{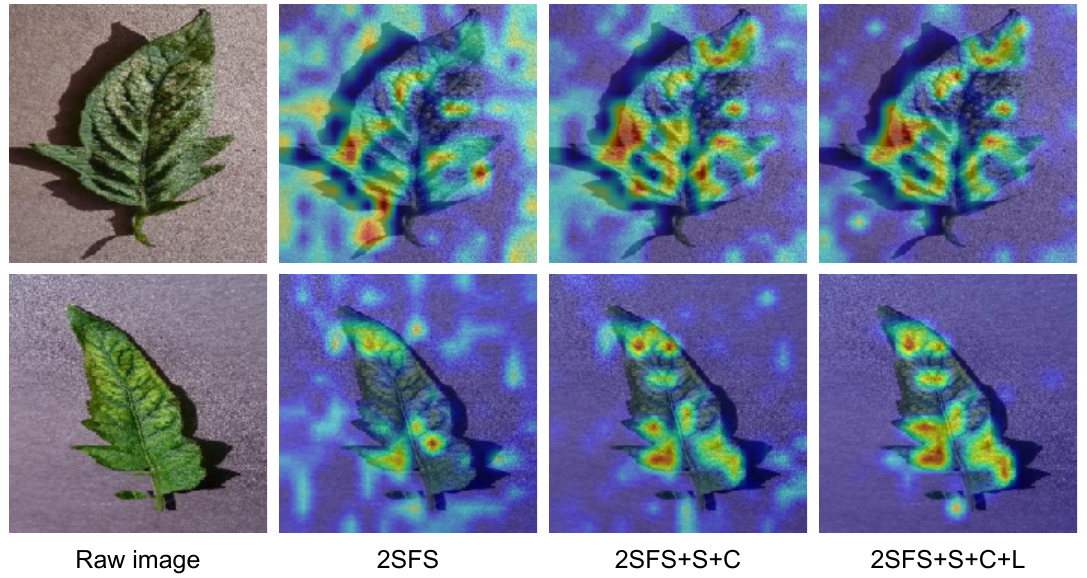}
\caption{Grad-CAM visualizations for the ground-truth class with progressive integration of Symptom and Comparative prototypes (S+C) and Local evidence calibration (L). Each row shows the raw image followed by the corresponding heatmaps from 2SFS, 2SFS+S+C, and the complete 2SFS-SCL.\label{fig:patch_class_relevance}}
\end{figure}

\FloatBarrier
\endgroup

\section{Discussion}

The proposed 2SFS-SCL framework demonstrates strong classification performance across three datasets, two backbones, and five shot settings. Under the ViT-L/14 16-shot setting, it achieves the best results across all four evaluation metrics on PlantVillage, Cassava, and Pear, with mean accuracies of 93.16\%, 79.30\%, and 96.56\%, respectively. These results validate the effectiveness of semantic enhancement and local evidence calibration for few-shot fine-grained plant disease classification. The performance improvements can be attributed to two main aspects:

First, description-guided class text representations provide more specific, comprehensive, and interpretable class semantics for fine-grained plant disease recognition. Methods such as CLIP-Adapter, Tip-Adapter-F, and 2SFS typically use class names and generic templates, which convey limited disease-specific diagnostic information. Prompt-learning methods such as CoOp and CoCoOp learn continuous prompt vectors from a few support samples, but such small support sets may not adequately cover typical visible symptoms, making it difficult to form accurate and comprehensive class text representations. The semantic basis of these learned vectors is also difficult to interpret directly. In contrast, 2SFS-SCL incorporates plant disease knowledge from outside the support set into class text representations through explicit symptom and comparative descriptions. Symptom descriptions summarize the typical visible symptoms of the target class, while comparative descriptions further highlight discriminative symptom cues for distinguishing similar classes. These descriptions also make the semantic basis of the representations easier to interpret and verify. The component ablation results further demonstrate that the enhanced class semantics complement limited visual supervision and improve few-shot classification performance.

Second, local evidence calibration mitigates the dilution of local diagnostic cues in global image–text matching. CLIP typically classifies images by matching global image representations with class text representations. However, diagnostic cues are often concentrated in local leaf regions and may be diluted by healthy tissue and background information in global representations. To address this limitation, 2SFS-SCL uses the enhanced class representations to aggregate class-relevant local evidence from patch tokens across multiple layers. The resulting local logits calibrate global logits through a residual formulation, supplementing global matching with local diagnostic cues. The component ablation results further demonstrate that this calibration mechanism improves few-shot fine-grained classification performance.

Beyond its strong in-domain performance, 2SFS-SCL also demonstrates good cross-domain generalization. In class-aligned cross-domain evaluation from PlantVillage to PlantDoc, it achieves the best overall classification performance across the three test splits and two backbones, outperforming 2SFS in all six settings. This advantage may stem from symptom semantics that remain applicable across domains. Specifically, diagnostic information conveyed by the descriptions, such as lesion color, shape, texture, and distribution, depends relatively little on specific image acquisition conditions and can provide a shared semantic reference for the same disease classes across domains. In addition, class-relevant local evidence may help reduce background interference in global image–text matching. In terms of adaptation, 2SFS-SCL first learns task-specific vision-language representations, then freezes the encoders and optimizes the classifier, preserving the representations used for cross-domain prediction while further refining in-domain decision boundaries. Comparisons between single-stage and two-stage adaptation under the same optimization budget further indicate that this strategy helps balance in-domain classification performance and cross-domain generalization.

Despite its promising performance, 2SFS-SCL still has several limitations. Under some low-shot settings, classification performance varies considerably across sampled support sets, highlighting the need for more stable few-shot adaptation. In addition, 2SFS-SCL is designed for single-label disease classification with predefined classes and does not yet support unknown disease recognition or the diagnosis of multiple co-occurring diseases.

\section{Conclusions}

This paper presented 2SFS-SCL, a description-guided two-stage CLIP adaptation method for few-shot fine-grained plant disease classification. By using symptom and comparative descriptions, the method incorporates plant disease knowledge from outside the support set into class text representations, making them more discriminative. Guided by the enhanced class representations, the model aggregates class-relevant local visual evidence to calibrate global image–text matching, supplementing it with local diagnostic cues that may be diluted in global representations. Furthermore, to balance task adaptation and cross-domain generalization, the method adopts a two-stage adaptation strategy: it first updates a small subset of encoder parameters to learn task-specific vision-language representations, then freezes the encoders and optimizes a linear classifier. Extensive experiments demonstrate that 2SFS-SCL achieves strong classification performance across multiple datasets, backbones, and shot settings, as well as good cross-domain generalization, confirming the effectiveness of the proposed method.

Future work could focus on improving the applicability of 2SFS-SCL in complex agricultural scenarios. Specifically, few-shot adaptation mechanisms could be refined to mitigate fluctuations in classification performance across sampled support sets. The method could also be extended to unknown disease recognition and the diagnosis of multiple co-occurring diseases to broaden its range of practical applications.

\supplementary{The supplementary file provides details of comparative description generation and selection, including the prompt template and representative examples.}

\authorcontributions{Conceptualization, Z.Y. and H.Z.; methodology, Z.Y.; software, Z.Y.; validation, Z.Y. and Q.Z.; formal analysis, Z.Y.; investigation, Z.Y.; resources, H.Z. and Q.Z.; data curation, Z.Y.; writing---original draft preparation, Z.Y.; writing---review and editing, H.Z. and Q.Z.; visualization, Z.Y.; supervision, H.Z.; project administration, H.Z.; funding acquisition, H.Z. All authors have read and agreed to the published version of the manuscript.}

\funding{This work was supported by the Central Guidance for Local Science and Technology Development Fund (Grant No. ZYYD2025QY19) and the China University Industry-Academia-Research Innovation Fund (Grant No. 2025XJS023).}

\institutionalreview{Not applicable.}

\informedconsent{Not applicable.}

\begingroup
\Urlmuskip=0mu plus 1mu
\dataavailability{PlantVillage, Cassava, and PlantDoc are publicly available. They can be downloaded from the following websites: \url{https://github.com/spMohanty/PlantVillage-Dataset}, \url{https://www.kaggle.com/c/cassava-disease}, and \url{https://github.com/pratikkayal/PlantDoc-Dataset}. For Pear, the authors collected 523 images in the field and obtained 1,151 from the Roboflow Universe dataset ``Pear Pests and Diseases'' (Version 1) at \url{https://universe.roboflow.com/yanjishun/pear_pests-and-diseases/dataset/1}. The field-acquired Pear images and the dataset splits are available from the corresponding author upon reasonable request.}
\endgroup

\acknowledgments{The authors acknowledge the guidance, technical assistance, and constructive feedback provided by their advisors and colleagues throughout this research. GPT-5.5 (OpenAI) was used to generate candidate comparative descriptions. ChatGPT (GPT-5.6 and GPT-6, OpenAI) was also used to assist with manuscript translation, language polishing, and proofreading. The authors have reviewed and edited the output and take full responsibility for the content of this publication.}

\conflictsofinterest{The authors declare no conflicts of interest.}

\begin{adjustwidth}{-\extralength}{0cm}

\reftitle{References}

\Urlmuskip=0mu plus 1mu
\bibliography{refs}

\PublishersNote{}
\end{adjustwidth}
\end{document}


\maketitle

These supplementary materials present the protocol for generating and selecting comparative descriptions and provide representative examples.

\section*{S1. Comparative Description Generation and Selection}

For PlantVillage, we retrieve the other classes most semantically similar to each target class based on cosine similarities between symptom prototypes, forming its similar-class set. For Cassava and Pear, which have fewer classes, we use all classes other than the target class as similar classes. Let \(N_s\) and \(G\) denote the number of similar classes and the number of candidate descriptions generated for each target--similar class pair, respectively. The class counts and generation settings for each dataset are summarized in Table~\ref{tab:generation_settings}.

\begingroup
\small
\centering
\setlength{\tabcolsep}{6pt}
\renewcommand{\arraystretch}{1.28}
\captionsetup{font=small,skip=6pt}
\captionof{table}{Dataset class counts and candidate generation settings.}
\label{tab:generation_settings}
\begin{tabular*}{\textwidth}{@{\extracolsep{\fill}}lccc@{}}
\toprule
\textbf{Dataset} & \textbf{Classes} & \textbf{Similar classes} ($N_s$) & \textbf{Candidates per pair} ($G$)\\
\midrule
PlantVillage & 38 & 10 & 10\\
Cassava & 5 & 4 & 25\\
Pear & 8 & 7 & 15\\
\bottomrule
\end{tabular*}
\par
\endgroup
\medskip

We then use GPT-5.5 to generate comparative descriptions for each target--similar class pair. With the similar class as a reference, the candidates describe visible diagnostic features of the target class in specific, concise phrases, avoiding negative or contrastive wording. The fixed prompt template for candidate generation is presented below with standardized terminology:
\begin{quote}
\small
Generate comparative visual descriptions for fine-grained plant disease classification.

For each target\_class vs similar\_class pair, describe visible features that make the target\_class visually distinguishable from the similar\_class. Use the similar\_class only for comparison, but output positive visual symptoms of the target\_class.

Each description will be inserted into:
``a photo of \{crop\} \{disease\} leaves with symptoms of \{description\}.''

Requirements:

1. Generate exactly \{num\_descriptions\_per\_pair\} descriptions for each similar\_class.

2. Each description should be a concrete visible symptom phrase, usually 8--17 words.

3. Mention only visual evidence.

4. Do not mention class names, pathogen names, mechanisms, treatment, climate, geography, or non-visual information.

5. Avoid contrastive or negative wording such as ``unlike'', ``rather than'', ``without'', ``not'', or ``no''.

6. Return valid JSON only.

\begin{minipage}{\linewidth}
Good examples:

- ``olive-brown velvety spots on leaf surfaces''

- ``irregular dark lesions with soft edges''

- ``clusters of dark specks merging into blotches''

- ``yellow halos surrounding small necrotic spots''

- ``powdery white coating across upper leaf surfaces''
\end{minipage}

\begin{minipage}{\linewidth}
Bad examples:

- ``olive-brown spots rather than clean green leaves''

- ``dark lesions without circular rings''

- ``yellowing unlike healthy foliage''

- ``symptoms of Apple scab''

- ``disease-specific visual patterns''
\end{minipage}

Output schema must be exactly one JSON object with target\_class and comparative\_descriptions. Preserve target\_class and similar\_class strings exactly as in the task file. Generate exactly the requested count per similar\_class. Return only valid JSON.
\end{quote}
The generation task uses the following input and output fields:
\begin{itemize}[itemsep=1pt,parsep=0pt,topsep=4pt]
\item The input fields are \texttt{target\_class}, \texttt{similar\_classes}, and \texttt{num\_descriptions\_per\_pair}.
\item The top-level output fields are \texttt{target\_class} and \texttt{comparative\_descriptions}.
\item Each entry in \texttt{comparative\_descriptions} contains a \texttt{similar\_class} field and a \texttt{descriptions} list containing the corresponding candidates.
\end{itemize}

We rank these candidates in descending order of the selection score defined in Section 2.4.2 of the main text and retain the top \(M\) descriptions for each class, with \(M=5\). Candidate generation and selection are performed offline before model adaptation. For each dataset--backbone combination, the same selected descriptions are used across different shot settings and their corresponding support sets.

\section*{S2. Representative Comparative Descriptions}

Examples of selected comparative descriptions under the ViT-B/16 setting are shown below.

\noindent\textbf{Apple scab (PlantVillage):}
\begin{itemize}[itemsep=1pt,parsep=0pt,topsep=4pt]
\item coalesced brown blotches creating ragged necrotic leaf areas
\item coalesced dark flecks creating ragged necrotic patches across leaf areas
\item coalescing brown spots creating ragged necrotic patches across leaf blades
\item tan necrotic spots with darker rims on puckered leaves
\item tan necrotic spots with darker rims spreading across broad leaf blades
\end{itemize}

\noindent\textbf{Cassava bacterial blight (Cassava):}
\begin{itemize}[itemsep=1pt,parsep=0pt,topsep=4pt]
\item angular water-soaked leaf spots with dark brown centers and yellow halos
\item irregular dead leaf areas with sharp brown edges and yellow halos
\item irregular rust-brown blotches with narrow yellow margins across broad leaves
\item water-soaked angular spots progressing into extensive brown leaf scorch
\item yellow halos around sunken chocolate-brown spots on flat leaf surfaces
\end{itemize}